\documentclass[11pt,a4paper]{article}
\usepackage[table]{xcolor}
\usepackage{times,latexsym}
\usepackage{url}
\usepackage[T1]{fontenc}
\usepackage{amsfonts}
\usepackage{amsmath}
\usepackage{todonotes}
\usepackage{booktabs}
\usepackage{adjustbox}
\usepackage{rotating}
\usepackage{layouts}
\usepackage{enumitem}%
\usepackage{afterpage}%
\usepackage{float}%
\usepackage{lipsum}%
\usepackage{tipa}

\definecolor{pos}{RGB}{76,144,186}
\definecolor{neg}{RGB}{222,102,62}
\definecolor{art}{RGB}{200,200,200}

\setlist[itemize]{noitemsep, topsep=0pt}

\usepackage[hang,flushmargin]{footmisc}

\usepackage[acceptedWithA]{tacl2018v2}

\usepackage{xspace,mfirstuc,tabulary}

\newcommand{\sys}{ABNIRML}

\newif\iftaclinstructions
\taclinstructionsfalse %
\iftaclinstructions
\renewcommand{\confidential}{}
\renewcommand{\anonsubtext}{(No author info supplied here, for consistency with
TACL-submission anonymization requirements)}
\newcommand{\instr}
\fi

\iftaclpubformat %

\else

\fi

\newcommand\smm[1]{#1}
\newcommand\smb[1]{#1}

\usepackage{array}
\newcolumntype{H}{>{\setbox0=\hbox\bgroup}c<{\egroup}@{}}

\title{\sys: Analyzing the Behavior of Neural IR Models\vspace{4pt}}

\author{
{\bf Sean MacAvaney}$^\mathbf{\dagger}$\thanks{\xspace \xspace Currently at the University of Glasgow. Work done in part during an internship at the Allen Institute for AI.} \qquad
Sergey Feldman$^\mathbf{\ddagger}$ \qquad
{\bf Nazli Goharian}$^\mathbf{\dagger}$ \\
{\bf Doug Downey}$^\mathbf{\ddagger}$ \qquad
{\bf Arman Cohan}$^\mathbf{\ddagger}$$^\mathbf{\S}$ \vspace{8pt} \\
  $^\mathbf{\dagger}$
 IR Lab, Georegetown University, Washington, DC \\
  $^\mathbf{\ddagger}$
 Allen Institute for AI, Seattle, WA \\
  $^\mathbf{\S}$
  Paul G. Allen School of Computer Science, University of Washington, WA\\
  {\tt\small \{sean,nazli\}@ir.cs.georgetown.edu} \\
  {\tt\small \{sergey,dougd,armanc\}@allenai.org}
}

\date{}

\begin{document}
\maketitle
\begin{abstract}
Pretrained contextualized language models such as BERT and T5 have established a new state-of-the-art for ad-hoc search. However, it is not yet well-understood why these methods are so effective, what makes some variants more effective than others, and what pitfalls they may have.  We present a new comprehensive framework for Analyzing
the Behavior of Neural IR ModeLs (\sys), which includes new types of diagnostic probes that allow us to test several characteristics---such as writing styles, factuality, sensitivity to paraphrasing and word order---that are not addressed by previous techniques.  To demonstrate the value of the framework, we conduct an extensive empirical study that yields insights into the factors that contribute to the neural model's gains, and identify potential unintended biases the models exhibit. 
Some of our results confirm conventional wisdom, like that recent neural ranking models rely less on exact term overlap with the query, and instead leverage richer linguistic information, evidenced by their higher sensitivity to word and sentence order. \smm{Other results are more surprising, such as that some models (e.g., T5 and ColBERT) are biased towards factually correct (rather than simply relevant) texts. Further, some characteristics vary even for the same base language model, and other characteristics can appear due to random variations during model training.\footnote{Code: \url{https://github.com/allenai/abnriml}}}

\end{abstract}

\section{Introduction}

Pre-trained
contextualized
language models
such as BERT~\cite{Devlin2019BERTPO} are state-of-the-art for a wide variety of natural language processing tasks~\cite{Xia2020WhichA}. In Information Retrieval (IR), these models have brought about large improvements in the task of {\em ad-hoc retrieval}---ranking documents by their relevance to a textual query~\cite{Lin2020PretrainedTF,Nogueira2019PassageRW,macavaney:sigir2019-cedr,Dai2019DeeperTU}---where the models increasingly dominate competition leaderboards ~\cite{craswell2020overview,dalton2019cast}.

Despite this success, little is understood about {\em why} pretrained language models are effective for ad-hoc ranking.
Previous work has shown that traditional IR axioms, e.g. that increased term frequency should correspond to higher relevance, do {\em not} explain the behavior of recent neural models \citep{Cmara2020DiagnosingBW}. Outside of IR, others have examined what characteristics contextualized language models learn in general \cite{Liu2019LinguisticKA, Rogers2020API, Loureiro2020LanguageMA}, but it remains unclear if these qualities are valuable for ad-hoc ranking specifically.
Thus, new approaches are necessary to characterize models.

We propose a new framework aimed at Analyzing the Behavior of Neural IR ModeLs (\sys\footnote{Pronounced /ab'n\textschwa{}rm\textschwa{}l/, similar to ``abnormal''.}), which aims to probe the sensitivity of ranking models on specific textual properties. Probes consist of samples comprised of a query and two contrastive documents. We propose three strategies for building probes. The ``measure and match'' strategy (akin to the diagnostic datasets proposed by~\citet{Rennings2019AnAA})
constructs probing samples by controlling one measurement (e.g., term frequency) and varying another (e.g., document length) using samples from an existing IR collection. \smm{Unlike~\citet{Rennings2019AnAA}, our framework generalizes the idea to any measurable characteristic, rather than relying chiefly on prior proposed IR axioms.} A second strategy, ``textual manipulation,'' probes the effect that altering the text of a document text has on its ranking. Finally, a ``dataset transfer'' strategy constructs probes from non-IR datasets.  The new probes allow us to isolate model characteristics---such as sensitivity to word order\smm{, degree of lexical simplicity, or even factuality}---that cannot be analyzed using other approaches. 

Using our new framework, we perform the first large-scale analysis of neural IR models.
We compare today's leading ranking techniques, including those using BERT~\cite{Devlin2019BERTPO} and T5~\cite{Raffel2019ExploringTL}, methods focused on efficiency like DocT5Query~\cite{Nogueira2020DocumentRW} and EPIC~\cite{macavaney:sigir2020-epic}, \smm{ and dense retrieval models like  ANCE~\cite{Xiong2021ApproximateNN} and ColBERT~\cite{Khattab2020ColBERTEA}}.\footnote{\smm{Although a multitude of other models exist, it is impractical to investigate them all. We instead focus on a representative sample of the recent and successful models and well-known baselines to provide context.}}
Some of our results establish widely-believed, but not-yet-verified conjectures about neural models.  For example, we show that neural models can exploit richer linguistic signals than classical term-matching metrics like BM25: when controlling for term frequency match, the neural models detect document relevance much more accurately than the BM25 baseline.  Similarly, unlike prior approaches, rankers based on BERT and T5 are heavily influenced by word order: shuffling the words in a document consistently lowers the document's score relative to the unmodified version, and neural rankers show a sensitivity to sentence order that is completely absent in classical models.  Other findings from \sys\ are more surprising. \smm{For example, we find that the T5 and ColBERT models we examine prefer answers that are factually correct, implying that they encode and utilize some real-world knowledge. Further, although this knowledge may be a result of the model's pre-training process, it is not necessarily utilized as a ranking signal, given that other models that use the same base language model do not have the same preference.}
Our battery of probes also uncover a variety of other findings, including that adding additional text to documents can often exhibit adverse behavior in neural models---decreasing the document's score when the added text is relevant, and increasing score when the added text is irrelevant.

In summary, we present a new framework (\sys) for performing analysis of ad-hoc ranking models. We then demonstrate how the framework can provide insights into ranking model characteristics by providing the most comprehensive analysis of neural ranking models to date.
Our software implementation of the framework is easily extensible, facilitating the replication of our results and further analyses in future work.

\section{\sys}

\begin{figure*}[t]
\centering
\includegraphics[scale=0.5]{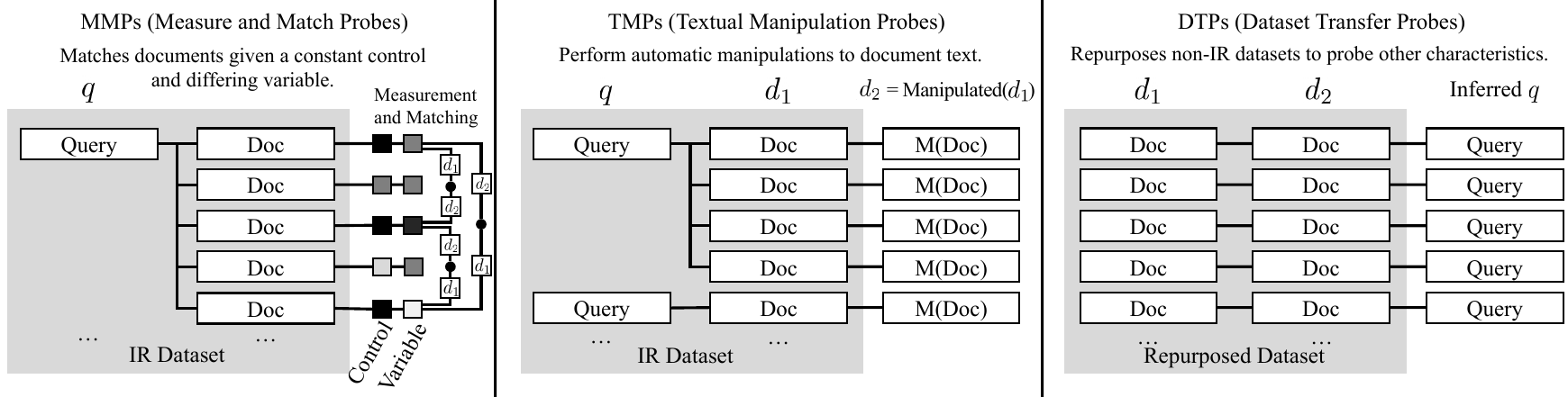}
\caption{Overview of strategies for constructing probes. Each probe in \sys{} is comprised of samples, each of which consists of a query ($q$) and two documents ($d_1$ and $d_2$).}
\label{fig:method}
\end{figure*}

In order to characterize the behavior of ranking models we construct several diagnostic probes. Each probe aims to evaluate specific properties of ranking models and probe their behavior (e.g., if they are heavily influenced by term matching, discourse and coherence, conciseness/verbosity, writing styles, etc). We formulate three different approaches to construct probes (Measure and Match, Textual Manipulation, and Dataset Transfer).

In ad-hoc ranking, a query (expressed in natural language) is submitted by a user to a search engine, and a ranking function provides the user with a list of natural language documents sorted by relevance to the query. More formally, let $R(q,d)\in\mathbb{R}$ be a ranking function, which maps a given query $q$ and document $d$ (each being a natural-language sequence of terms) to a real-valued ranking score. At query time, documents in a collection $D$ are scored using $R(\cdot)$ for a given query $q$, and ranked by the scores (conventionally, sorted descending by score). 
Learning-to-rank models optimize a set of parameters for the task of relevance ranking based on training data. %

\subsection{Document Pair Probing}

We utilize a \textit{document pair probing} strategy, in which probes are comprised of samples, each of which consists of a query and two documents that differ primarily in some characteristic of interest (e.g., succinctness). The ranking scores of the two documents are then compared (with respect to the query). This allows the isolation of particular model preferences. For instance, a probe could consist of summarized and full texts of news articles; models that consistently rank summaries over full texts prefer succinct text.

More formally, each document pair probe consists of a collection of samples $S$, where each $\langle q,d_1,d_2\rangle \in S$ is a 3-tuple consisting of a query (or query-like text, $q$), and two documents (or document-like texts, $d_1$ and $d_2$). The relationship between $d_1$ and $d_2$ (with respect to $q$) for each sample defines the probe. E.g., a probe testing summarization could be defined as: (1) $d_2$ is a summary of $d_1$, and (2) $d_1$ is relevant to query $q$.  %

Almost all of our probes are {\em directional}, where $d_2$ has some attribute that $d_1$ lacks, and we measure the effect of this attribute on ranking.  Specifically, each sample in the probe is scored as: ($+1$) scoring $d_1$ above $d_2$ (a \textit{positive} effect), ($-1$) scoring $d_2$ above $d_1$ (a \textit{negative} effect), or ($0$) a \textit{neutral} effect. Formally, the effect $\mathit{eff}(\cdot)$ of a given sample is defined as:
\begin{equation}
\mathit{eff}(q,d_1,d_2)=
{\scriptstyle
\begin{cases}
\scriptstyle \phantom{-}1  & \scriptstyle R(q,d_1) - R(q,d_2) > \delta \\
\scriptstyle -1 & \scriptstyle R(q,d_1) - R(q,d_2) < -\delta \\
\scriptstyle \phantom{-}0  & \scriptstyle -\delta \leq R(q,d_1) - R(q,d_2) \leq \delta \\
\end{cases}
}
\end{equation}
The parameter $\delta$ adjusts how large the score difference between the scores of $d_1$ and $d_2$ must be in order to count as positive or negative effect. This allows us to disregard small changes to the score that are unlikely to affect the final ranking. In practice, $\delta$ depends on the ranking model because each model scores on different scales. Therefore we tune $\delta$ for each model (see Section~\ref{sec:delta}).

{\em Symmetric} probes are different from directional ones in that $d_1$ and $d_2$ are exchangeable; for example, we experiment with one symmetric probe in which $d_1$ and $d_2$ are paraphrases of each other.  For symmetric probes only the magnitude of score difference is meaningful, and thus $\mathit{eff}$ outputs 1 if the absolute value of the difference is larger than $\delta$, and 0 otherwise.

A model's performance on a particular probe is summarized by a single score $s$ that averages the effect of all samples in the probe:
\begin{equation}
\scalebox{0.9}{
$s = \frac{1}{|S|} \sum\limits_{\langle q,d_1,d_2\rangle\in S} \mathit{eff}(q,d_1,d_2)$
}
\end{equation}
Note that this score is in the interval $[-1,1]$ for directional probes \smm{and $[0,1]$ for symmetric probes. For directional probes,} positive scores indicate a stronger preference towards documents from group 1 ($d_1$ documents), and negative scores indicate a preference towards documents from group 2 ($d_2$ documents). Scores near 0 indicate no strong preference or preferences that are split roughly evenly; disentangling these two cases requires analyzing individual effect scores.

There are several important differences between our setup and the ``diagnostic dataset'' approach proposed by \citet{Rennings2019AnAA}. First, by including the $\delta$ threshold, we ensure that our probes measure differences that can affect the final order in ranked lists. Second, by including the ``neutral effect'' case in our scoring function, we  distinguish between cases in which $d_1$ or $d_2$ are preferred and cases where neither document is strongly preferred. And finally, our probes are aimed at describing model behavior, rather than evaluating models.
\smm{For instance, one of our tests measures whether the model prefers succinct or elaborative text---whether this preference is desirable depends on the application or even the particular user.}

\subsection{Document Pair Probing Strategies}

In this work, we explore three strategies for designing document pair probes.  As discussed below, the strategies have different strengths and weaknesses. When used in concert, they allow us to characterize a wide variety of model behaviors. 
Figure~\ref{fig:method} provides an overview of the strategies.

\subsubsection{Measure and Match Probes (MMPs)}

Some surface-level characteristics of documents, such as its Term Frequency (TF) for a given query, are both easy to measure and valuable for characterizing models.  Comparing the ranking scores of two documents that differ in one characteristic but are otherwise similar yields evidence of how the characteristic influences model behavior. Measure and Match Probes (MMPs) follow such an approach.  MMPs involve first measuring the characteristics of judged query-document pairs in an IR dataset. Then, the pairs are matched to form probe samples consisting of a {\em control} (a characteristic that approximately matches between the documents, such as document length), and a {\em variable} (which differs between documents, such as TF).  %
Probes employed in previous work to verify existing ranking axioms~\cite{Cmara2020DiagnosingBW,Rennings2019AnAA}\footnote{An example is TFC1 from~\cite{Fang2004AFS}, which suggests that higher TFs should be mapped to higher scores.} are instances of MMPs. %

For our experiments, we design MMPs to explore the relationship between the primary IR objective (document relevance) and the classical IR ranking signal (TF, potentially controlling for document length).  We are motivated to explore this relationship because TF has long been used as a core signal for ranking algorithms; a departure from monotonically increasing the score of a document as TF increases would represent a fundamental shift in the notion of relevance scoring~\cite{Fang2004AFS}.
Specifically, we explore the following characteristics in MMPs:
\begin{itemize}
\item \textbf{Relevance}: the human-assessed graded relevance score of a document to the given query.
\item \textbf{Length}: the document length, in total number of non-stopword tokens.
\item \textbf{TF}: the individual Porter-stemmed Term Frequencies of non-stopword terms from the query. To determine when the TF of two documents are different, we use the condition that the TF of at least one query term in $d_1$ must be greater than the same term in $d_2$, and that no term in $d_1$ can have a lower TF than the corresponding term in $d_2$.
\item \textbf{Overlap}: the proportion of non-stopword terms in the document that appear in the query. Put another way, the total TF divided by the document length.
\end{itemize}

Each of these characteristics is used as both a variable (matching based on differing values) and a control (matching based on identical values). In our experiments, we examine all pairs of these characteristics, greatly expanding upon IR axioms investigated in prior work.

\smm{We note that the MMPs that we explore in this work do not cover all prior IR axioms. For instance, axioms SMTC1--3, proposed by~\citet{Fang2006SemanticTM}, suggest behaviors related to the occurrence of semantically-similar terms. Although MMPs can be constructed to test these, we assert that other types of probes are more suitable to testing these behaviors. We test textual fluency, formality, and simplicity (all of which are specific types of semantic similarity) while controlling for the meaning of the text using dataset transfer probes (Section~\ref{sec:dtt}).}

\subsubsection{Textual Manipulation Probes (TMPs)}

Not all characteristics are easily captured with MMPs. For instance, it would be difficult to probe the sensitivity to word order with MMPs; it is unlikely to find naturally-occurring document pairs that use the same words but in a different order. \smm{Nevertheless, it is valuable to understand the extent to which models are affected by characteristics like this, given that traditional bag-of-words models are unaffected by word order and that there is evidence that word order is unimportant when fine-tuning recent neural models~\cite{Sinha2021MaskedLM,Alleman2021SyntacticPR}.} %
To overcome these limitations, we propose Textual Manipulation Probes (TMPs). TMPs apply a manipulation function to scored documents from an existing IR dataset. For example, for probing word order, we can use a simple manipulation function that, given a document $d_1$, creates a corresponding synthetic document $d_2$ by shuffling the order of the words in each sentence.  The degree to which a model prefers $d_1$ is then a measure of its preference for proper word order.  %
Prior works that use a similar approach for probing ranking methods include the collection perturbation tests of \citet{Fang2011DiagnosticEO} (which perform operations like removing documents from the collection and deleting individual terms from documents) and a diagnostic dataset proposed by~\citet{Rennings2019AnAA} (which tests the effect of duplicating the document: an adaptation of a traditional ranking axiom). Although TMPs allow probing a wider variety of characteristics than MMPs, we note that they involve constructing artificial data; $d_2$ may not resemble documents seen in practice. Despite this, their versatility make TMPs an attractive choice for a variety of characteristics.

We now detail the specific TMPs we explore in our experiments.  We use TMPs to verify a key difference we expect to hold between neural models and previous rankers: because neural models are pretrained on large bodies of running text, they should make better use of richer linguistic features like word order.  We investigate this with TMPs that {\bf shuffle words} in the document.  We also probe which aspects of word order are important, through TMPs that only shuffle a small number of non-content words ({\bf prepositions}) and TMPs that only shuffle the {\bf sentence order}, but not the individual words within each sentence.  Further, another important distinction of pretrained neural models is that they process unaltered text, without classical normalization like {\bf stopword removal} or {\bf lemmatization}; we introduce TMPs that study these manipulations.\footnote{We use SpaCy's~\cite{spacy2} lemmatizer, rather than e.g. a stemmer, because the outputs from a stemming function like Porter are often not found in the lexicon of models like BERT.} \smm{Recognizing changes such as lemmatization and word shuffling can drastically alter the text, we also include a more subtle TMP that applies typical typograhpical errors (\textbf{typos}) by replacing words with common misspellings.\footnote{We use this list of common errors in English text: \url{https://en.wikipedia.org/wiki/Commonly_misspelled_English_words}}} We also evaluate the recent, effective technique of using neural models to add content ({\bf DocT5Query terms} \cite{Nogueira2020DocumentRW}) to each document to aid IR, and contrast this with a complementary TMP that adds a {\bf non-relevant sentence} to the document.

\subsubsection{Dataset Transfer Probes (DTPs)}\label{sec:dtt}
Even with MMPs and TMPs, some characteristics may still be difficult to measure. For instance, for attributes like textual {\em fluency} (the degree to which language sounds like a native speaker wrote it), we would need to find pairs of otherwise-similar documents with measurable differences in fluency (for an MMP) or identify ways to automatically manipulate fluency (for a TMP), both of which would be difficult. %
To probe characteristics like these, we propose Dataset Transfer Probes (DTPs). In this setting, a dataset built for a purpose other than ranking is repurposed to probe a ranking model's behavior. For example, one could create a DTP from a dataset of human-written textual fluency pairs (e.g., from the JFLEG dataset~\cite{napoles-sakaguchi-tetreault:2017:EACLshort}) to sidestep challenges in both measurement and manipulation. Text pair datasets are abundant, allowing us to probe a wide variety of characteristics, like fluency, formality, and succinctness. With these probes, $d_1$ and $d_2$ can be easily defined by the source dataset. In some cases, external information can be used to infer a corresponding $q$, such as using the title of the article as a query for news article summarization tasks, a technique that has been studied before to train ranking models~\cite{macavaney:sigir2019-nyt}. In other cases, queries can be artificially generated, \smm{as long as the text resembles a likely query.} %

We first use DTPs to investigate the important question of whether models exhibit confounding preferences for {\em stylistic} features of text are at least partially independent of relevance.  \smm{Specifically, we first investigate paraphrases in general, and then move on to check the specific qualities of} fluency, formality, simplicity, lexical bias, and succinctness. \smm{We then use DTPs to test the capacity of models to encode and utilize real-world knowledge through probes that measure a model's tendency to select factual answers.}

\smm{The TMPs described in the previous section probe the sensitivity of models to word order. In this case, the {\em words} remain the same, but {\em meaning} is \smb{altered}. It is natural to wonder whether model behaviors would be similar if the meaning is preserved when using different words. This motivates a {\bf paraphrase} DTP.
We construct this probe from the Microsoft Paraphrase Corpus (MSPC).\footnote{\url{https://www.microsoft.com/en-us/download/details.aspx?id=52398}} We select $d_1$ and $d_2$ from all text pairs labeled as paraphrases. Note that this is the first example of a symmetric probe, as there is no directionality in the paraphrase relation; the assignment of $d_1$ and $d_2$ is arbitrary. We generate $q$ by randomly selecting a noun chunk that appears in both versions of the text, ensuring a query that is relevant to both texts. (If no such chunk exists, we discard the sample.) By selecting a noun chunk, the query remains reasonably similar to a real query.}

\smm{Although the paraphrase probe can tell us whether models distinguish between text with similar meaning, it cannot tell us what characteristics it favors when making such a distinction. To gain insights here, we propose several directional probes based on stylistic differences that result in similar meanings. One such characteristic is textual {\bf fluency}.} We propose a DTP using the JFLEG dataset~\cite{napoles-sakaguchi-tetreault:2017:EACLshort}. This dataset contains sentences from English-language fluency tests. Each non-fluent sentence is corrected for fluency by four fluent English speakers to make the text sound `natural' (changes include grammar and word usage changes). We treat each fluent text as a $d_1$ paired with the non-fluent $d_2$, and use the strategy used for paraphrases to generate $q$.

We probe {\bf formality} by building a DTP from the GYAFC dataset~\cite{Rao2018DearSO}. This dataset selects sentences from Yahoo Answers and has four annotators make edits to the text that either improve the formality (for text that is informal), or reduce the formality (for text that is already formal). We treat formal text as $d_1$ and informal text as $d_2$. Since the text came from Yahoo Answers, we can link the text back to the original questions using the Yahoo L6 dataset.\footnote{\url{https://webscope.sandbox.yahoo.com/catalog.php?datatype=l\&did=11}} We treat the question (title) as $q$. In cases where we cannot find the original text or there are no overlapping non-stopword lemmas from $q$ in both $d_1$ and $d_2$, we discard the sample.

\smm{The \textbf{simplicity} of text indicates the ease of reading a particular text. We test the effect of \textit{lexical} text simplicity using the WikiTurk dataset provided by \citet{xu-etal-2016-optimizing}. In this dataset, sentences from Wikipedia were edited to make them simpler by Amazon Turk workers. We treat the simplified text as $d_1$, the original text as $d_2$, and we use the query construction technique from the paraphrase probe for $q$.}

\smm{Text can also express similar ideas but with differing degrees of subjectivity or bias. We construct a \textbf{neutrality} DTP using the Wikipedia Neutrality Corpus (WNC) dataset~\cite{Pryzant2020AutomaticallyNS}.  This corpus consists of sentences that were corrected by Wikipedia editors to enforce the platform's neutral point of view. We use the neutral text as $d_1$, the biased text as $d_2$, and we use the query construction technique from the paraphrase probe for $q$.}

\smm{An idea can also be expressed in greater or lesser detail.} To probe whether models have a preference for \textbf{succinctness}, we construct DTPs from summarization datasets, using the assumption that a document's summary will be more succinct than its full text. We utilize two datasets to conduct this probe: XSum~\cite{xsum-emnlp}, and CNN/DailyMail~\cite{See2017GetTT}. The former uses extremely concise summaries from BBC articles, usually consisting of a single sentence. The CNN/DailyMail dataset uses slightly longer bullet point list summaries, usually consisting of around 3 sentences. For these probes, we use the title of the article as $q$, the summarized text as $d_1$, and the article body as $d_2$. When there is no overlap between the non-stopword lemmas of $q$ in both $d_1$ and $d_2$, we discard the samples. We further sub-sample the dataset at 10\% because the datasets are already rather large. To handle the long full text in BERT and EPIC, we use the passage aggregation strategy proposed by \citet{macavaney:sigir2019-cedr}.

\smm{Moving beyond probes that express similar ideas, we explore the extent to which models are aware of real-world knowledge using a \textbf{factuality} probe. This probe is motivated by the intuition that contextualized language models may be memorizing facts from the pre-training corpus when determining relevance. We construct this probe from the Natural Questions~\cite{Kwiatkowski2019NQ} dataset. We make use of the known answer text from NQ by replacing it with a similar answer. Similar answers must be of the same entity type\footnote{Entities extracted using SpaCy: Person (PER), Location (LOC), Geo-Political Entity (GPE), Nationality/Religion/etc. (NORP), or Organization (ORG).} and have the same number of non-stopword tokens. We discard samples where the question text contains the answer text (e.g., this-or-that questions). We use the factual text as $d_1$, the non-factual version as $d_2$, and the question text as $q$. Note that this probe can be considered both a DTP and a TMP. We decide to consider it to primarily be a DTP because it makes use of data specific to this external dataset (i.e., answer strings).}

\section{Experimental Setup}

\subsection{Datasets}

We use the MS-MARCO passage dataset~\cite{Campos2016MSMA} to train the neural ranking models. The training subset contains approximately $809k$ natural-language questions from a query log (with an average length of 7.5 terms) and $8.8$ million candidate answer passages (with an average length of 73.1 terms). Due to its scale in number of queries, it is shallowly annotated, almost always containing fewer than 3 positive judgments per query. This dataset is frequently used for training neural ranking models. \smm{Importantly, it also has been shown to effectively transfer relevance signals to other collections~\cite{nogueiradoc2query}, making it suitable for use with DTPs, which may include text from other domains.}

We build MMPs and TMPs using the TREC Deep Learning 2019 passage dataset~\cite{craswell2020overview} \smm{and the ANTIQUE passage ranking dataset~\cite{Hashemi2020Antique}}. TREC DL uses the MS-MARCO passage collection and has 43 queries with deep relevance judgments (on average, 215 per query). The judgments are graded as highly relevant (7\%), relevant (19\%), topical (17\%), and non-relevant (56\%), allowing us to make more fine-grained comparisons. \smm{We use the test subset of ANTIQUE, which contains 200 queries with 33 judgments per query. These judgments are graded as convincing (20\%), possibly correct (18\%), on-topic (37\%), and off-topic (25\%).} We opt to perform our analysis in a passage ranking setting to eliminate effects of long document aggregation\smm{---which is challenging for some neural models given a maximum sequence length in the underlying model---}given that this is an area with many model varieties that is still under active investigation~\cite{li:arxiv2020-parade}.

\subsection{Models}

We compare a sample of several models covering a traditional lexical model (BM25), a conventional learning-to-rank approach (LightGBM), and neural models based on contextualized language models. We include two models that focus on query-time computational efficiency, and two representative models that use dense retrieval. The neural models represent a sample of the recent state-of-the-art ranking models. \smm{For each model, we provide the MRR (minimum relevance of 2) performance on the TREC DL 2019 passage benchmark when re-ranking the provided candidate passages.}

\textbf{BM25}.
We use the Terrier~\cite{ounis06terrier-osir} implementation of BM25 with default parameters. BM25 is an unsupervised model that incorporates the lexical features of term frequency (TF), inverse document frequency (IDF), and document length. (TREC DL 2019 MRR: 0.627.)

\smb{\textbf{WMD}. As a second unsupervised model, we use the Word Mover's Distance~\cite{Kusner2015FromWE} over (non-contextualized) GloVe~\cite{Pennington2014GloVeGV} embeddings (\texttt{glove-wiki-gigaword-100}). We use the implementation from the Gensim~\cite{rehurek2011gensim} Python package.
(TREC DL 2019 MRR: 0.364.)}

\smb{\textbf{SBERT}. As an unsupervised model based on a contextualized language model, we use SBERT's~\cite{reimers-2019-sentence-bert} pre-trained Bi-encoder model, trained on Semantic Textual Similarity, Natural Language Inference, and Quora Duplicate Question Detection data in multiple languages.\footnote{\texttt{distilbert-multilingual-nli-stsb-quora-ranking}} This approach has been shown by \citet{Litschko2021EvaluatingMT} to be able to effectively perform cross-lingual retrieval.
(TREC DL 2019 MRR: 0.465.)}

\textbf{LGBM}~\cite{Ke2017LightGBMAH}. As a non-neural learning-to-rank baseline, we use the Light Gradient Boosting Machine model currently used by the Semantic Scholar search engine~\cite{feldman_2020}.\footnote{\url{https://github.com/allenai/s2search}} This public model was trained on clickthrough data from this search engine, meaning that it services various information needs (e.g., navigational and topical queries). Not all of the model's features are available in our setting (e.g., recency, in-links, etc.), so we only supply the text-based features like lexical overlap and scores from a light-weight language model~\cite{heafield-etal-2013-scalable}.
(TREC DL 2019 MRR: 0.580.)

\textbf{VBERT}~\cite{Devlin2019BERTPO}. We use a BERT model, which uses a linear ranking layer atop a BERT pretrained transformer language model~\cite{Nogueira2019PassageRW,macavaney:sigir2019-cedr,Dai2019DeeperTU}. (This setup goes by several names in the literature, including Vanilla BERT (VBERT), monoBERT, BERT-CAT, etc.) We fine-tune the \texttt{bert-base-uncased} model for this task using the official training sequence of the MS-MARCO passage ranking dataset. (TREC DL 2019 MRR: 0.809.)

\textbf{T5}~\cite{Raffel2019ExploringTL}.
The Text-To-Text Transformer ranking model~\cite{nogueiradoc2query} scores documents by predicting whether the concatenated query, document, and control tokens is likely to generate the term `true' or `false'. We use the models released by the authors, which were tuned on the MS-MARCO passage ranking dataset. \smm{We test both the \texttt{t5-base} (T5-B) and \texttt{t5-large} (T5-L) models to gain insights into the effect of model size.} (TREC DL 2019 MRR: 0.868 (T5-B), 0.857 (T5-L).)

\textbf{EPIC}~\cite{macavaney:sigir2020-epic}. This is an efficiency-focused BERT-based model, which separately encodes query and document content into vectors that are the size of the source lexicon (where each element represents the importance of the corresponding term in the query/document). We use the \texttt{bert-base-uncased} model, and tune the model for ranking using the train split of the MS-MARCO passage ranking dataset with the code released by the EPIC authors with default settings. (TREC DL 2019 MRR: 0.809.)

\textbf{DT5Q}~\cite{nogueiradoc2query}. The T5 variant of the Doc2Query model (DT5Q) generates additional terms to add to a document using a T5 model. The expanded document can be efficiently indexed, boosting the weight of terms likely to match queries. We use the model released by the authors, which was trained using the MS-MARCO passage training dataset. For our probes, we generate four queries to add to each document. As was done in the original paper, we use BM25 as a scoring function over the expanded documents. (TREC DL 2019 MRR: 0.692.)

\smm{\textbf{ANCE}~\cite{Xiong2021ApproximateNN}. This is a representation-based dense retrieval model that is trained using a contrastive learning technique. It is designed for single-stage dense retrieval. We use the model weights released by the original authors, which is based on the RoBERTa~\cite{Liu2019RoBERTaAR} base model. (TREC DL 2019 MRR: 0.852.)}

\smm{\textbf{ColBERT}~\cite{Khattab2020ColBERTEA}. This is a two-stage dense retrieval approach that uses multiple representations for each document (one per WordPiece token). It makes use of both a first-stage approximate nearest neighbor search to find candidate documents and a re-ranking stage to calculate the precise ranking scores. It is based on the \texttt{bert-base-uncased} model. We use the model weights released by the original authors. (TREC DL 2019 MRR: 0.873.)}

\begin{table*}
\centering\small
\renewcommand{\arraystretch}{0.9}

\setlength{\tabcolsep}{5.7pt}
\scalebox{0.75}{
\begin{tabular}{@{}llrrrrrrrrrrrr@{}}

\toprule
Variable & Control & BM25 & WMD & SBERT & LGBM & DT5Q & VBERT & EPIC & T5-B & T5-L & ColBERT & ANCE & Samples \\
\midrule
\multicolumn{2}{l}{\bf TREC DL 2019} \\ \midrule
Relevance & Length & \cellcolor{pos!20}$+0.40$ & \cellcolor{pos!13}$+0.27$ & \cellcolor{pos!21}$+0.43$ & \cellcolor{pos!19}$+0.40$ & \cellcolor{pos!23}$+0.48$ & \cellcolor{pos!29}$+0.58$ & \cellcolor{pos!26}$+0.54$ & \cellcolor{pos!30}$+0.61$ & \cellcolor{pos!33}$+0.66$ & \cellcolor{pos!30}$+0.61$ & \cellcolor{pos!26}$+0.53$ & 19676 \\
 & TF & \cellcolor{neg!1}$-0.03$ & \cellcolor{pos!5}$+0.11$ & \cellcolor{pos!12}$+0.25$ & \cellcolor{pos!1}$+0.04$ & \cellcolor{pos!4}$+0.10$ & \cellcolor{pos!16}$+0.34$ & \cellcolor{pos!13}$+0.27$ & \cellcolor{pos!21}$+0.43$ & \cellcolor{pos!26}$+0.53$ & \cellcolor{pos!23}$+0.47$ & \cellcolor{pos!22}$+0.45$ & 31619 \\
 & Overlap & \cellcolor{pos!20}$+0.41$ & \cellcolor{pos!7}$+0.15$ & \cellcolor{pos!19}$+0.39$ & \cellcolor{pos!17}$+0.34$ & \cellcolor{pos!23}$+0.47$ & \cellcolor{pos!27}$+0.55$ & \cellcolor{pos!24}$+0.50$ & \cellcolor{pos!30}$+0.61$ & \cellcolor{pos!32}$+0.65$ & \cellcolor{pos!29}$+0.60$ & \cellcolor{pos!24}$+0.49$ & 4762 \\
\midrule
Length & Relevance & \cellcolor{neg!2}$-0.05$ & \cellcolor{neg!5}$-0.10$ & \cellcolor{neg!0}$-0.01$ & \cellcolor{pos!2}$+0.04$ & \cellcolor{neg!3}$-0.07$ & \cellcolor{neg!0}$^*$$-0.01$ & \cellcolor{neg!3}$-0.08$ & \cellcolor{pos!0}$+0.01$ & \cellcolor{pos!0}$+0.00$ & \cellcolor{pos!0}$^*$$+0.00$ & \cellcolor{pos!0}$+0.00$ & 515401 \\
 & TF & \cellcolor{neg!6}$-0.14$ & \cellcolor{neg!4}$-0.08$ & \cellcolor{pos!0}$^*$$+0.02$ & \cellcolor{pos!1}$+0.02$ & \cellcolor{neg!4}$-0.09$ & \cellcolor{neg!4}$-0.09$ & \cellcolor{neg!7}$-0.15$ & \cellcolor{pos!0}$+0.01$ & \cellcolor{neg!0}$^*$$-0.00$ & \cellcolor{pos!1}$^*$$+0.03$ & \cellcolor{pos!2}$+0.06$ & 88582 \\
 & Overlap & \cellcolor{pos!25}$+0.51$ & \cellcolor{pos!0}$^*$$+0.02$ & \cellcolor{pos!7}$+0.15$ & \cellcolor{pos!13}$+0.26$ & \cellcolor{pos!11}$+0.24$ & \cellcolor{pos!9}$+0.20$ & \cellcolor{pos!5}$+0.11$ & \cellcolor{pos!9}$+0.19$ & \cellcolor{pos!9}$+0.18$ & \cellcolor{pos!8}$+0.18$ & \cellcolor{pos!7}$+0.15$ & 3963 \\
\midrule
TF & Relevance & \cellcolor{pos!44}$+0.88$ & \cellcolor{pos!24}$+0.49$ & \cellcolor{pos!17}$+0.34$ & \cellcolor{pos!24}$+0.50$ & \cellcolor{pos!36}$+0.73$ & \cellcolor{pos!20}$+0.41$ & \cellcolor{pos!23}$+0.48$ & \cellcolor{pos!19}$+0.38$ & \cellcolor{pos!20}$+0.42$ & \cellcolor{pos!19}$+0.39$ & \cellcolor{pos!17}$+0.35$ & 303058 \\
 & Length & \cellcolor{pos!49}$+1.00$ & \cellcolor{pos!32}$+0.65$ & \cellcolor{pos!23}$+0.46$ & \cellcolor{pos!29}$+0.59$ & \cellcolor{pos!42}$+0.84$ & \cellcolor{pos!27}$+0.54$ & \cellcolor{pos!30}$+0.61$ & \cellcolor{pos!25}$+0.51$ & \cellcolor{pos!26}$+0.53$ & \cellcolor{pos!26}$+0.53$ & \cellcolor{pos!23}$+0.47$ & 19770 \\
 & Overlap & \cellcolor{pos!39}$+0.79$ & \cellcolor{pos!0}$^*$$+0.02$ & \cellcolor{pos!8}$+0.18$ & \cellcolor{pos!18}$+0.37$ & \cellcolor{pos!17}$+0.36$ & \cellcolor{pos!12}$+0.26$ & \cellcolor{pos!8}$+0.17$ & \cellcolor{pos!13}$+0.26$ & \cellcolor{pos!11}$+0.24$ & \cellcolor{pos!12}$+0.25$ & \cellcolor{pos!9}$+0.19$ & 2294 \\
\midrule
Overlap & Relevance & \cellcolor{pos!34}$+0.70$ & \cellcolor{pos!23}$+0.47$ & \cellcolor{pos!10}$+0.22$ & \cellcolor{pos!9}$+0.20$ & \cellcolor{pos!26}$+0.52$ & \cellcolor{pos!9}$+0.19$ & \cellcolor{pos!12}$+0.25$ & \cellcolor{pos!8}$+0.17$ & \cellcolor{pos!8}$+0.18$ & \cellcolor{pos!7}$+0.14$ & \cellcolor{pos!9}$+0.18$ & 357470 \\
 & Length & \cellcolor{pos!37}$+0.75$ & \cellcolor{pos!29}$+0.59$ & \cellcolor{pos!16}$+0.32$ & \cellcolor{pos!17}$+0.35$ & \cellcolor{pos!29}$+0.59$ & \cellcolor{pos!15}$+0.31$ & \cellcolor{pos!17}$+0.35$ & \cellcolor{pos!13}$+0.28$ & \cellcolor{pos!14}$+0.29$ & \cellcolor{pos!13}$+0.27$ & \cellcolor{pos!14}$+0.30$ & 20819 \\
 & TF & \cellcolor{pos!43}$+0.88$ & \cellcolor{pos!12}$+0.25$ & \cellcolor{neg!0}$^*$$-0.00$ & \cellcolor{neg!1}$-0.03$ & \cellcolor{pos!23}$+0.47$ & \cellcolor{pos!5}$+0.11$ & \cellcolor{pos!8}$+0.17$ & \cellcolor{pos!1}$+0.04$ & \cellcolor{pos!3}$+0.06$ & \cellcolor{pos!1}$^*$$+0.03$ & \cellcolor{pos!2}$+0.04$ & 13980 \\
\midrule\multicolumn{2}{l}{\bf ANTIQUE} \\ \midrule
Relevance & Length & \cellcolor{neg!8}$-0.17$ & \cellcolor{neg!4}$^*$$-0.09$ & \cellcolor{pos!5}$+0.12$ & \cellcolor{neg!7}$-0.15$ & \cellcolor{neg!4}$-0.09$ & \cellcolor{pos!11}$+0.23$ & \cellcolor{neg!0}$^*$$-0.01$ & \cellcolor{pos!12}$+0.26$ & \cellcolor{pos!17}$+0.35$ & \cellcolor{pos!6}$+0.13$ & \cellcolor{pos!12}$+0.24$ & 2257 \\
 & TF & \cellcolor{neg!3}$-0.07$ & \cellcolor{pos!0}$^*$$+0.01$ & \cellcolor{pos!8}$+0.18$ & \cellcolor{pos!1}$^*$$+0.02$ & \cellcolor{pos!2}$+0.04$ & \cellcolor{pos!11}$+0.23$ & \cellcolor{pos!11}$+0.23$ & \cellcolor{pos!17}$+0.34$ & \cellcolor{pos!23}$+0.46$ & \cellcolor{pos!14}$+0.28$ & \cellcolor{pos!16}$+0.33$ & 5586 \\
 & Overlap & \cellcolor{neg!0}$^*$$-0.01$ & \cellcolor{pos!0}$^*$$+0.00$ & \cellcolor{pos!13}$+0.26$ & \cellcolor{pos!1}$^*$$+0.03$ & \cellcolor{pos!5}$+0.12$ & \cellcolor{pos!19}$+0.39$ & \cellcolor{pos!8}$+0.16$ & \cellcolor{pos!20}$+0.42$ & \cellcolor{pos!23}$+0.47$ & \cellcolor{pos!15}$+0.31$ & \cellcolor{pos!17}$+0.36$ & 1211 \\
\midrule
Length & Relevance & \cellcolor{pos!2}$+0.04$ & \cellcolor{neg!3}$^*$$-0.07$ & \cellcolor{pos!6}$+0.13$ & \cellcolor{pos!11}$+0.23$ & \cellcolor{pos!0}$+0.02$ & \cellcolor{neg!3}$-0.07$ & \cellcolor{pos!11}$+0.22$ & \cellcolor{pos!6}$+0.12$ & \cellcolor{pos!8}$+0.17$ & \cellcolor{pos!8}$+0.17$ & \cellcolor{pos!11}$+0.23$ & 36164 \\
 & TF & \cellcolor{neg!23}$-0.47$ & \cellcolor{neg!4}$^*$$-0.09$ & \cellcolor{pos!5}$+0.12$ & \cellcolor{pos!1}$+0.04$ & \cellcolor{neg!11}$-0.23$ & \cellcolor{neg!6}$-0.13$ & \cellcolor{pos!12}$+0.25$ & \cellcolor{pos!1}$+0.03$ & \cellcolor{pos!9}$+0.19$ & \cellcolor{pos!7}$+0.15$ & \cellcolor{pos!11}$+0.24$ & 8296 \\
 & Overlap & \cellcolor{pos!33}$+0.67$ & \cellcolor{pos!3}$^*$$+0.07$ & \cellcolor{pos!8}$+0.17$ & \cellcolor{pos!16}$+0.33$ & \cellcolor{pos!17}$+0.34$ & \cellcolor{pos!1}$^*$$+0.04$ & \cellcolor{pos!17}$+0.35$ & \cellcolor{pos!6}$+0.12$ & \cellcolor{pos!8}$+0.17$ & \cellcolor{pos!10}$+0.21$ & \cellcolor{pos!13}$+0.28$ & 902 \\
\midrule
TF & Relevance & \cellcolor{pos!34}$+0.69$ & \cellcolor{pos!11}$^*$$+0.23$ & \cellcolor{pos!18}$+0.37$ & \cellcolor{pos!28}$+0.57$ & \cellcolor{pos!27}$+0.56$ & \cellcolor{pos!12}$+0.24$ & \cellcolor{pos!26}$+0.53$ & \cellcolor{pos!19}$+0.38$ & \cellcolor{pos!21}$+0.42$ & \cellcolor{pos!23}$+0.46$ & \cellcolor{pos!22}$+0.45$ & 19900 \\
 & Length & \cellcolor{pos!49}$+1.00$ & \cellcolor{pos!23}$^*$$+0.48$ & \cellcolor{pos!25}$+0.50$ & \cellcolor{pos!33}$+0.68$ & \cellcolor{pos!42}$+0.84$ & \cellcolor{pos!19}$+0.39$ & \cellcolor{pos!29}$+0.59$ & \cellcolor{pos!18}$+0.36$ & \cellcolor{pos!15}$+0.31$ & \cellcolor{pos!27}$+0.55$ & \cellcolor{pos!19}$+0.40$ & 1397 \\
 & Overlap & \cellcolor{pos!46}$+0.92$ & \cellcolor{pos!2}$^*$$+0.06$ & \cellcolor{pos!7}$+0.14$ & \cellcolor{pos!17}$+0.36$ & \cellcolor{pos!22}$+0.45$ & \cellcolor{pos!4}$^*$$+0.08$ & \cellcolor{pos!17}$+0.35$ & \cellcolor{pos!7}$+0.15$ & \cellcolor{pos!10}$+0.22$ & \cellcolor{pos!12}$+0.25$ & \cellcolor{pos!14}$+0.28$ & 553 \\
\midrule
Overlap & Relevance & \cellcolor{pos!20}$+0.42$ & \cellcolor{pos!14}$^*$$+0.29$ & \cellcolor{pos!4}$+0.09$ & \cellcolor{pos!0}$+0.01$ & \cellcolor{pos!17}$+0.35$ & \cellcolor{pos!10}$+0.21$ & \cellcolor{pos!0}$^*$$+0.01$ & \cellcolor{pos!3}$+0.07$ & \cellcolor{pos!1}$^*$$+0.03$ & \cellcolor{pos!2}$+0.04$ & \cellcolor{neg!0}$^*$$-0.01$ & 27539 \\
 & Length & \cellcolor{pos!33}$+0.67$ & \cellcolor{pos!16}$^*$$+0.33$ & \cellcolor{pos!10}$+0.22$ & \cellcolor{pos!17}$+0.35$ & \cellcolor{pos!24}$+0.48$ & \cellcolor{pos!5}$+0.10$ & \cellcolor{pos!12}$+0.25$ & \cellcolor{pos!6}$+0.13$ & \cellcolor{pos!4}$^*$$+0.08$ & \cellcolor{pos!9}$+0.20$ & \cellcolor{pos!9}$+0.18$ & 1224 \\
 & TF & \cellcolor{pos!43}$+0.87$ & \cellcolor{pos!10}$^*$$+0.21$ & \cellcolor{pos!3}$+0.07$ & \cellcolor{neg!2}$-0.05$ & \cellcolor{pos!21}$+0.44$ & \cellcolor{pos!6}$+0.14$ & \cellcolor{neg!6}$-0.13$ & \cellcolor{neg!0}$-0.01$ & \cellcolor{neg!3}$-0.07$ & \cellcolor{neg!0}$^*$$-0.00$ & \cellcolor{neg!3}$-0.08$ & 4498 \\
\bottomrule

\end{tabular}
}

\caption{Results of Measure and Match Probes (MMPs) on the TREC DL 2019 \smm{and ANTIQUE datasets}. Positive scores indicate a preference towards a higher value of the variable. Scores marked with * are not statistically significant (see Section~\ref{sec:sig}).}
\label{tab:mmt}
\end{table*}

\subsection{Choosing $\delta$}\label{sec:delta}

Recall that $\delta$ indicates the minimum absolute difference of scores in a document pair probe to have a positive or negative effect. Since each model scores documents on a different scale, we empirically choose a $\delta$ per model. \smm{We do this by re-ranking the official set from TREC DL 2019.} Among the top 10 results, we calculate the differences between each adjacent pair of scores (i.e., $\{R(q,d_1)-R(q,d_2), R(q,d_2)-R(q,d_3), ..., R(q,d_9)-R(q,d_{10})\}$, where $d_i$ is the $i$th highest scored document for $q$). We set $\delta$ to the median difference. By setting the threshold this way, we can expect the differences captured by the probes to have an effect on the final ranking score \textit{at least} half the time. \smm{We explore this further in Section~\ref{sec:delta_effect}. \smm{Note that choosing a constant $\delta$ over one that is assigned per-query allows for testing probes where a complete corpus is not available, as is the case for some DTPs.}}

\subsection{Significance Testing}\label{sec:sig}

We use a two-sided paired T-Test to determine the significance (pairs of $R(q,d_1)$ and $R(q,d_2)$). We use a Bonferroni correction over each table to correct for multiple tests, and test for $p<0.01$.

\subsection{Software and Libraries}

We use the following software to conduct our experiments: PyTerrier~\cite{MacDonald2020DeclarativeEI}, OpenNIR~\cite{macavaney:wsdm2020-onir}, \texttt{ir\_datasets}~\cite{macavaney:sigir2021-irds}, Transformers~\cite{Wolf2019HuggingFacesTS}, sentence-transformers~\cite{reimers-2019-sentence-bert}, Anserini~\cite{Yang2018AnseriniRR}\smb{, and Gensim~\cite{rehurek2011gensim}.}

{
\renewcommand{\arraystretch}{0.9}
\begin{table*}[t]
\setlength{\tabcolsep}{5pt}
\centering\small

\scalebox{0.75}{
\begin{tabular}{@{}llrrrrrrrrrrrr@{}}

\toprule
Probe & Dataset & BM25 & WMD & SBERT & LGBM & DT5Q & VBERT & EPIC & T5-B & T5-L & ColBERT & ANCE & Samples \\
\midrule
Rem. Stops/Punct & DL19 & \cellcolor{pos!0}$^*$$+0.00$ & \cellcolor{neg!4}$^*$$-0.09$ & \cellcolor{neg!11}$-0.23$ & \cellcolor{neg!9}$-0.20$ & \cellcolor{neg!2}$-0.04$ & \cellcolor{pos!8}$+0.18$ & \cellcolor{neg!39}$-0.78$ & \cellcolor{neg!37}$-0.74$ & \cellcolor{neg!39}$-0.80$ & \cellcolor{neg!34}$-0.68$ & \cellcolor{neg!29}$-0.59$ & 9259 \\
 & ANT & \cellcolor{pos!1}$^*$$+0.04$ & \cellcolor{neg!9}$^*$$-0.19$ & \cellcolor{neg!19}$-0.38$ & \cellcolor{neg!11}$-0.24$ & \cellcolor{neg!3}$-0.07$ & \cellcolor{neg!12}$-0.25$ & \cellcolor{neg!39}$-0.78$ & \cellcolor{neg!31}$-0.64$ & \cellcolor{neg!40}$-0.81$ & \cellcolor{neg!37}$-0.74$ & \cellcolor{neg!35}$-0.70$ & 6540 \\
\midrule
Lemmatize & DL19 & \cellcolor{pos!0}$+0.00$ & \cellcolor{neg!8}$-0.18$ & \cellcolor{pos!2}$+0.05$ & \cellcolor{neg!0}$-0.02$ & \cellcolor{pos!0}$^*$$+0.01$ & \cellcolor{neg!2}$-0.04$ & \cellcolor{neg!12}$-0.25$ & \cellcolor{neg!20}$-0.42$ & \cellcolor{neg!22}$-0.44$ & \cellcolor{neg!18}$-0.38$ & \cellcolor{neg!15}$-0.31$ & 9259 \\
 & ANT & \cellcolor{pos!2}$+0.04$ & \cellcolor{neg!0}$^*$$-0.01$ & \cellcolor{neg!1}$-0.04$ & \cellcolor{neg!4}$-0.09$ & \cellcolor{pos!0}$+0.00$ & \cellcolor{neg!10}$-0.22$ & \cellcolor{neg!12}$-0.25$ & \cellcolor{neg!14}$-0.30$ & \cellcolor{neg!23}$-0.47$ & \cellcolor{neg!12}$-0.25$ & \cellcolor{neg!15}$-0.31$ & 6392 \\
\midrule
Shuf. Words & DL19 & \cellcolor{pos!0}$^*$$+0.00$ & \cellcolor{neg!10}$-0.21$ & \cellcolor{neg!2}$-0.06$ & \cellcolor{neg!12}$-0.25$ & \cellcolor{neg!5}$-0.11$ & \cellcolor{neg!18}$-0.38$ & \cellcolor{neg!38}$-0.76$ & \cellcolor{neg!32}$-0.65$ & \cellcolor{neg!37}$-0.76$ & \cellcolor{neg!38}$-0.76$ & \cellcolor{neg!19}$-0.40$ & 9260 \\
 & ANT & \cellcolor{pos!1}$^*$$+0.04$ & \cellcolor{neg!5}$^*$$-0.11$ & \cellcolor{neg!5}$-0.10$ & \cellcolor{neg!12}$-0.25$ & \cellcolor{neg!6}$-0.13$ & \cellcolor{neg!30}$-0.61$ & \cellcolor{neg!33}$-0.67$ & \cellcolor{neg!32}$-0.65$ & \cellcolor{neg!37}$-0.75$ & \cellcolor{neg!33}$-0.67$ & \cellcolor{neg!29}$-0.58$ & 6545 \\
\midrule
Shuf. Sents. & DL19 & \cellcolor{neg!0}$^*$$-0.00$ & \cellcolor{neg!0}$^*$$-0.01$ & \cellcolor{neg!3}$-0.06$ & \cellcolor{neg!0}$^*$$-0.00$ & \cellcolor{neg!1}$^*$$-0.02$ & \cellcolor{neg!6}$-0.13$ & \cellcolor{neg!9}$-0.19$ & \cellcolor{neg!9}$-0.20$ & \cellcolor{neg!6}$-0.14$ & \cellcolor{neg!7}$-0.14$ & \cellcolor{neg!5}$-0.10$ & 7290 \\
 & ANT & \cellcolor{pos!0}$^*$$-0.00$ & \cellcolor{neg!0}$^*$$-0.02$ & \cellcolor{neg!1}$-0.04$ & \cellcolor{neg!0}$^*$$-0.00$ & \cellcolor{neg!1}$^*$$-0.02$ & \cellcolor{neg!8}$-0.17$ & \cellcolor{neg!9}$-0.20$ & \cellcolor{neg!11}$-0.22$ & \cellcolor{neg!11}$-0.22$ & \cellcolor{neg!6}$-0.13$ & \cellcolor{neg!7}$-0.14$ & 4211 \\
\midrule
Shuf. Prepositions & DL19 & \cellcolor{pos!0}$+0.01$ & \cellcolor{neg!10}$-0.21$ & \cellcolor{neg!1}$-0.02$ & \cellcolor{neg!0}$-0.02$ & \cellcolor{pos!0}$^*$$+0.02$ & \cellcolor{neg!0}$-0.01$ & \cellcolor{neg!5}$-0.11$ & \cellcolor{neg!14}$-0.28$ & \cellcolor{neg!15}$-0.31$ & \cellcolor{neg!9}$-0.18$ & \cellcolor{neg!11}$-0.24$ & 9239 \\
 & ANT & \cellcolor{pos!2}$+0.05$ & \cellcolor{neg!5}$^*$$-0.11$ & \cellcolor{neg!1}$-0.04$ & \cellcolor{neg!1}$-0.03$ & \cellcolor{pos!0}$+0.01$ & \cellcolor{neg!6}$-0.12$ & \cellcolor{neg!7}$-0.16$ & \cellcolor{neg!15}$-0.30$ & \cellcolor{neg!17}$-0.36$ & \cellcolor{neg!8}$-0.18$ & \cellcolor{neg!14}$-0.29$ & 6186 \\
\midrule
Typos & DL19 & \cellcolor{neg!11}$-0.23$ & \cellcolor{neg!8}$-0.17$ & \cellcolor{pos!3}$^*$$+0.07$ & \cellcolor{neg!7}$-0.15$ & \cellcolor{neg!8}$-0.18$ & \cellcolor{neg!4}$-0.09$ & \cellcolor{neg!25}$-0.50$ & \cellcolor{neg!18}$-0.37$ & \cellcolor{neg!13}$-0.27$ & \cellcolor{neg!20}$-0.42$ & \cellcolor{neg!9}$-0.20$ & 8982 \\
 & ANT & \cellcolor{neg!16}$-0.32$ & \cellcolor{neg!20}$^*$$-0.41$ & \cellcolor{neg!4}$-0.09$ & \cellcolor{neg!13}$-0.27$ & \cellcolor{neg!13}$-0.27$ & \cellcolor{neg!20}$-0.40$ & \cellcolor{neg!22}$-0.45$ & \cellcolor{neg!19}$-0.38$ & \cellcolor{neg!19}$-0.40$ & \cellcolor{neg!27}$-0.56$ & \cellcolor{neg!17}$-0.36$ & 5551 \\
\midrule
+ DocT5Query & DL19 & \cellcolor{pos!16}$+0.34$ & \cellcolor{pos!22}$+0.45$ & \cellcolor{pos!16}$+0.33$ & \cellcolor{pos!20}$+0.41$ & \cellcolor{pos!7}$+0.15$ & \cellcolor{neg!11}$-0.22$ & \cellcolor{neg!31}$-0.63$ & \cellcolor{neg!27}$-0.54$ & \cellcolor{neg!29}$-0.60$ & \cellcolor{neg!24}$-0.50$ & \cellcolor{neg!23}$-0.47$ & 9260 \\
 & ANT & \cellcolor{pos!16}$+0.34$ & \cellcolor{pos!7}$^*$$+0.14$ & \cellcolor{pos!8}$+0.17$ & \cellcolor{pos!16}$+0.32$ & \cellcolor{pos!1}$^*$$+0.03$ & \cellcolor{neg!21}$-0.42$ & \cellcolor{neg!6}$-0.13$ & \cellcolor{neg!33}$-0.67$ & \cellcolor{neg!34}$-0.68$ & \cellcolor{neg!5}$^*$$-0.10$ & \cellcolor{neg!18}$-0.37$ & 6589 \\
\midrule
+ Non-Rel Sent. & DL19 & \cellcolor{neg!1}$-0.03$ & \cellcolor{neg!4}$-0.10$ & \cellcolor{pos!17}$+0.34$ & \cellcolor{pos!10}$+0.20$ & \cellcolor{pos!1}$+0.04$ & \cellcolor{pos!12}$+0.26$ & \cellcolor{pos!5}$+0.11$ & \cellcolor{pos!16}$+0.33$ & \cellcolor{pos!13}$+0.27$ & \cellcolor{pos!16}$+0.33$ & \cellcolor{pos!19}$+0.39$ & 9260 \\
 & ANT & \cellcolor{pos!12}$+0.25$ & \cellcolor{pos!1}$^*$$+0.04$ & \cellcolor{pos!19}$+0.38$ & \cellcolor{pos!15}$+0.31$ & \cellcolor{pos!12}$+0.25$ & \cellcolor{pos!4}$+0.08$ & \cellcolor{pos!23}$+0.47$ & \cellcolor{pos!13}$+0.28$ & \cellcolor{pos!14}$+0.30$ & \cellcolor{pos!16}$+0.34$ & \cellcolor{pos!20}$+0.41$ & 6346 \\
\bottomrule

\end{tabular}
}

\caption{Results of Text Manipulation Probes (TMPs) on the TREC DL 2019 \smm{and ANTIQUE datasets}. Positive scores indicate a preference for the manipulated document text; negative scores prefer the original text. Scores marked with * are not statistically significant (see Section~\ref{sec:sig}).
}
\label{tab:tmt}
\end{table*}
}

\section{Results \& Analysis}

\smm{We present results for MMPs in Table~\ref{tab:mmt}, TMPs in Table~\ref{tab:tmt}, and DTPs in Table~\ref{tab:dtt} and highlight our key findings in the order they appear in the tables.}

\smm{\textbf{Contextualized language models can distinguish relevance grades when TF is held constant.} From Table~\ref{tab:mmt}, we see that \smb{SBERT}, VBERT, EPIC, T5, ColBERT, and ANCE all are able to distinguish relevance when term frequency is constant with at least a score of \smb{+0.18} across both datasets. \smb{Perhaps surprisingly, this is even true for our transfer SBERT model, which is not trained on relevance ranking data. These results are in contrast with} models that score lexically (BM25, LGBM, and DT5Q), which score at most +0.10. The contextualized language models also perform better at distinguishing relevance grades than the other models when length and overlap are held constant, though by a lesser margin. When controlling for model type, it appears that the model's size is related to its effectiveness in this setting: the large version of T5 (T5-L, +0.53) performs better the base model (T5-B, +0.43).}

\smm{\textbf{Models generally have similar sensitivity to document length, TF, and overlap on TREC DL 2019.} With the exception of models that use BM25 for scoring (BM25 and DT5Q), all the models we explore have similar behaviors when varying length, TF, and overlap. This suggests that although signals like TF are not \textit{required} for EPIC, BERT, and T5 to rank effectively, they still remain an important signal when available. There are bigger differences between models when exploring the ANTIQUE dataset, suggesting differences in the models' capacity to generalize. We note that some of the largest differences relate to the relevance measurement, highlighting the differences in label definitions between the two datasets.}

\smm{\textbf{\smb{Trained} Contextualized language models are adversely affected by heavily-destructive pre-processing steps.} From Table~\ref{tab:tmt}, we find that removing stopwords and punctuation, performing lemmatization, and shuffling words negatively impacts most models across both datasets. \smb{Perhaps} this is expected, given that this text is dissimilar to the text \smb{the models were} pre-trained on. \smb{However, we note that the transfer SBERT model is far less affected by these operations, suggesting that these characteristics are not intrinsic to the contextualized language models, but rather a consequence of training them for relevance ranking.}
To gain further insights into the importance of word order, we control for local word order by only shuffling sentence order. We see that an effect remains for the contextualized models, though it is substantially reduced. This suggests that discourse-level signals (e.g., what topics are discussed earlier in a document) have some effect on the models, or the models encode some positional bias (e.g., preferring answers at the start of documents).
To understand if the word usage of particular terms is important (rather than overall coherence), we also try shuffling only the prepositions in the sentence. We find that this has an effect on some models (most notably, both T5 models and ANCE), but not other models, suggesting that some end up learning that although these terms have meaning in the text, they are often unimportant when it comes to ranking.}

\begin{table*}
\centering\small
\renewcommand{\arraystretch}{0.9}

\scalebox{0.7}{
\begin{tabular}{@{}llrrrrrrrrrrrr@{}}
\toprule
Probe & Dataset & BM25 & WMD & SBERT & LGBM & DT5Q & VBERT & EPIC & T5-B & T5-L & ColBERT & ANCE & Samples \\
\midrule
Paraphrase & MSPC & \cellcolor{pos!29}$0.60$ & \cellcolor{pos!44}$^*$$0.89$ & \cellcolor{pos!46}$0.94$ & \cellcolor{pos!0}$^*$$0.00$ & \cellcolor{pos!38}$0.76$ & \cellcolor{pos!40}$0.82$ & \cellcolor{pos!32}$0.65$ & \cellcolor{pos!45}$0.91$ & \cellcolor{pos!44}$0.88$ & \cellcolor{pos!42}$0.85$ & \cellcolor{pos!44}$0.90$ & 3421 \\
\midrule
Fluency & JFLEG & \cellcolor{pos!1}$+0.03$ & \cellcolor{neg!3}$^*$$-0.07$ & \cellcolor{pos!0}$^*$$+0.00$ & \cellcolor{pos!0}$^*$$-0.00$ & \cellcolor{pos!0}$^*$$+0.02$ & \cellcolor{pos!4}$+0.10$ & \cellcolor{pos!11}$+0.22$ & \cellcolor{pos!6}$+0.14$ & \cellcolor{pos!3}$+0.07$ & \cellcolor{pos!12}$+0.24$ & \cellcolor{pos!8}$+0.17$ & 5073 \\
 & (spellchecked) & \cellcolor{pos!0}$^*$$+0.01$ & \cellcolor{pos!2}$^*$$+0.05$ & \cellcolor{neg!1}$^*$$-0.03$ & \cellcolor{pos!0}$^*$$-0.00$ & \cellcolor{neg!0}$^*$$-0.01$ & \cellcolor{pos!3}$+0.07$ & \cellcolor{pos!9}$+0.20$ & \cellcolor{pos!6}$+0.14$ & \cellcolor{pos!6}$+0.13$ & \cellcolor{pos!9}$+0.18$ & \cellcolor{pos!4}$+0.09$ & 5187 \\
\midrule
Formality & GYAFC & \cellcolor{neg!1}$-0.03$ & \cellcolor{neg!1}$^*$$-0.03$ & \cellcolor{neg!7}$-0.15$ & \cellcolor{neg!4}$-0.09$ & \cellcolor{neg!3}$-0.07$ & \cellcolor{neg!2}$-0.05$ & \cellcolor{pos!7}$+0.16$ & \cellcolor{pos!0}$^*$$+0.01$ & \cellcolor{pos!7}$+0.15$ & \cellcolor{neg!3}$-0.07$ & \cellcolor{pos!2}$^*$$+0.05$ & 6721 \\
 &  - entertain. & \cellcolor{pos!1}$+0.04$ & \cellcolor{pos!0}$^*$$+0.01$ & \cellcolor{neg!5}$^*$$-0.11$ & \cellcolor{neg!1}$-0.04$ & \cellcolor{neg!0}$-0.01$ & \cellcolor{pos!2}$^*$$+0.04$ & \cellcolor{pos!9}$+0.19$ & \cellcolor{pos!5}$^*$$+0.11$ & \cellcolor{pos!11}$+0.23$ & \cellcolor{pos!0}$^*$$+0.01$ & \cellcolor{pos!3}$+0.08$ & 2960 \\
 &  - family & \cellcolor{neg!4}$-0.08$ & \cellcolor{neg!2}$^*$$-0.05$ & \cellcolor{neg!9}$-0.18$ & \cellcolor{neg!6}$-0.13$ & \cellcolor{neg!5}$-0.11$ & \cellcolor{neg!6}$-0.12$ & \cellcolor{pos!6}$^*$$+0.13$ & \cellcolor{neg!3}$-0.08$ & \cellcolor{pos!3}$^*$$+0.08$ & \cellcolor{neg!6}$-0.14$ & \cellcolor{pos!1}$^*$$+0.03$ & 3761 \\
\midrule
Simplicity & WikiTurk & \cellcolor{pos!6}$+0.13$ & \cellcolor{pos!10}$^*$$+0.21$ & \cellcolor{pos!3}$+0.07$ & \cellcolor{pos!0}$^*$$-0.00$ & \cellcolor{pos!2}$+0.05$ & \cellcolor{neg!1}$^*$$-0.03$ & \cellcolor{neg!0}$^*$$-0.01$ & \cellcolor{neg!3}$-0.08$ & \cellcolor{neg!6}$-0.13$ & \cellcolor{pos!0}$+0.01$ & \cellcolor{neg!1}$^*$$-0.03$ & 17849 \\
\midrule
Neutrality & WNC & \cellcolor{pos!15}$+0.31$ & \cellcolor{pos!17}$^*$$+0.34$ & \cellcolor{pos!5}$+0.11$ & \cellcolor{pos!0}$^*$$+0.00$ & \cellcolor{pos!6}$+0.13$ & \cellcolor{pos!5}$+0.11$ & \cellcolor{pos!3}$+0.07$ & \cellcolor{neg!0}$-0.00$ & \cellcolor{pos!1}$+0.03$ & \cellcolor{pos!6}$+0.13$ & \cellcolor{neg!0}$-0.00$ & 178252 \\
\midrule
Succinctness & XSum & \cellcolor{pos!33}$+0.66$ & \cellcolor{pos!45}$^*$$+0.91$ & \cellcolor{pos!29}$+0.58$ & \cellcolor{pos!9}$+0.18$ & \cellcolor{pos!32}$+0.66$ & \cellcolor{pos!24}$+0.49$ & \cellcolor{pos!9}$+0.18$ & \cellcolor{neg!4}$-0.09$ & \cellcolor{pos!3}$+0.07$ & \cellcolor{pos!16}$+0.33$ & \cellcolor{pos!23}$+0.47$ & 17938 \\
 & CNN & \cellcolor{pos!18}$+0.37$ & \cellcolor{pos!37}$^*$$+0.74$ & \cellcolor{pos!0}$^*$$+0.02$ & \cellcolor{neg!21}$-0.43$ & \cellcolor{pos!20}$+0.41$ & \cellcolor{pos!8}$+0.16$ & \cellcolor{neg!36}$-0.72$ & \cellcolor{neg!29}$-0.58$ & \cellcolor{neg!26}$-0.54$ & \cellcolor{neg!16}$-0.33$ & \cellcolor{neg!14}$-0.28$ & 7154 \\
 & Daily Mail & \cellcolor{neg!0}$^*$$-0.01$ & \cellcolor{pos!27}$+0.54$ & \cellcolor{neg!18}$-0.37$ & \cellcolor{neg!39}$-0.80$ & \cellcolor{pos!3}$+0.06$ & \cellcolor{neg!13}$-0.26$ & \cellcolor{neg!46}$-0.93$ & \cellcolor{neg!31}$-0.63$ & \cellcolor{neg!29}$-0.58$ & \cellcolor{neg!35}$-0.71$ & \cellcolor{neg!27}$-0.56$ & 18930 \\
\midrule
Factuality & NQ: PER & \cellcolor{neg!0}$^*$$-0.00$ & \cellcolor{pos!8}$+0.16$ & \cellcolor{neg!1}$-0.02$ & \cellcolor{neg!0}$-0.00$ & \cellcolor{neg!0}$-0.02$ & \cellcolor{neg!1}$^*$$-0.02$ & \cellcolor{neg!3}$-0.07$ & \cellcolor{pos!5}$+0.10$ & \cellcolor{pos!7}$+0.14$ & \cellcolor{pos!1}$+0.04$ & \cellcolor{pos!1}$+0.04$ & 72983 \\
 & NQ: GPE & \cellcolor{neg!0}$^*$$-0.00$ & \cellcolor{pos!11}$+0.22$ & \cellcolor{pos!0}$+0.02$ & \cellcolor{pos!0}$+0.00$ & \cellcolor{pos!0}$^*$$+0.00$ & \cellcolor{pos!4}$+0.09$ & \cellcolor{pos!0}$+0.00$ & \cellcolor{pos!13}$+0.27$ & \cellcolor{pos!15}$+0.30$ & \cellcolor{pos!11}$+0.22$ & \cellcolor{pos!6}$+0.12$ & 33528 \\
 & NQ: LOC & \cellcolor{neg!1}$^*$$-0.03$ & \cellcolor{pos!10}$+0.21$ & \cellcolor{neg!5}$^*$$-0.12$ & \cellcolor{neg!0}$^*$$-0.02$ & \cellcolor{neg!1}$^*$$-0.02$ & \cellcolor{pos!0}$^*$$+0.01$ & \cellcolor{pos!1}$^*$$+0.02$ & \cellcolor{pos!14}$+0.28$ & \cellcolor{pos!14}$+0.29$ & \cellcolor{pos!7}$+0.14$ & \cellcolor{pos!5}$^*$$+0.11$ & 962 \\
 & NQ: NORP & \cellcolor{pos!1}$+0.02$ & \cellcolor{pos!14}$+0.30$ & \cellcolor{pos!0}$+0.01$ & \cellcolor{pos!0}$+0.01$ & \cellcolor{pos!1}$+0.03$ & \cellcolor{pos!3}$+0.07$ & \cellcolor{pos!3}$+0.07$ & \cellcolor{pos!12}$+0.25$ & \cellcolor{pos!16}$+0.33$ & \cellcolor{pos!13}$+0.26$ & \cellcolor{pos!4}$+0.10$ & 4250 \\
 & NQ: ORG & \cellcolor{pos!0}$+0.01$ & \cellcolor{pos!16}$+0.34$ & \cellcolor{pos!0}$+0.01$ & \cellcolor{pos!0}$^*$$+0.00$ & \cellcolor{pos!0}$^*$$+0.01$ & \cellcolor{pos!3}$+0.07$ & \cellcolor{neg!0}$-0.01$ & \cellcolor{pos!16}$+0.33$ & \cellcolor{pos!19}$+0.38$ & \cellcolor{pos!9}$+0.19$ & \cellcolor{pos!6}$+0.13$ & 13831 \\
\bottomrule
\end{tabular}
}

\caption{Results of Dataset Transfer Probes (DTPs). The paraphrase probe is unsigned, as it is symmetric. Positive scores indicate a preference for fluent, formal, \smm{simplified, neutral (non-biased), succinct, and factual} text. Scores marked with * are not statistically significant (see Section~\ref{sec:sig}).}
\label{tab:dtt}
\end{table*}

\smm{\textbf{Lexical models handle typographical errors better \smb{than trained contextualized language models}.} In all but one case (ANCE DL19), BM25, LGBM, and DT5Q are negatively affected by typographical errors less than the \smb{trained} contextualized language models. This is a surprising result, given that contextualized language models should be able to learn common misspellings and treat them similarly to the original words \smb{(the transfer SBERT model largely ignores typos)}. This problem is particularly apparent for EPIC and ColBERT, which perform matching on the WordPiece level.}

\textbf{\smb{Trained} Contextualized models behave unexpectedly when additional content is introduced in documents.} 
We find that models that rely heavily on unigram matching (e.g., BM25) \smb{and the transfer SBERT model} respond positively to the addition of DocT5Query terms. Even the DocT5Query model itself sees an additional boost,
suggesting that weighting the expansion terms higher in the document may further improve the effectiveness of this model. However, the contextualized models often respond \textit{negatively} to these additions.
We also find that adding non-relevant sentences to the end of relevant documents often increases the ranking score of contextualized models. This is in contrast with models like BM25, in which the scores of relevant documents decrease with the addition of non-relevant information. From the variable length MMPs, we know that this increase in score is likely not due to increasing the length alone. Such characteristics may pose a risk to ranking systems based on contextualized models, in which content sources could aim to increase their ranking simply by adding non-relevant content to their documents.

\begin{figure*}
\centering
\scriptsize
\begin{tabular}{ccc}
(a) MMP: Dataset=DL19, V=Rel, C=Len &
(b) TMP: Rem. Stops/Punct Dataset=DL19 &
(c) DTT: Paraphrase Dataset=MSPC \\
\includegraphics[scale=0.5]{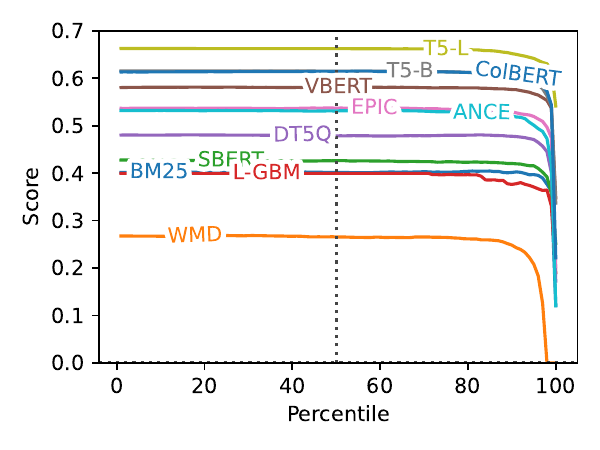} &
\includegraphics[scale=0.5]{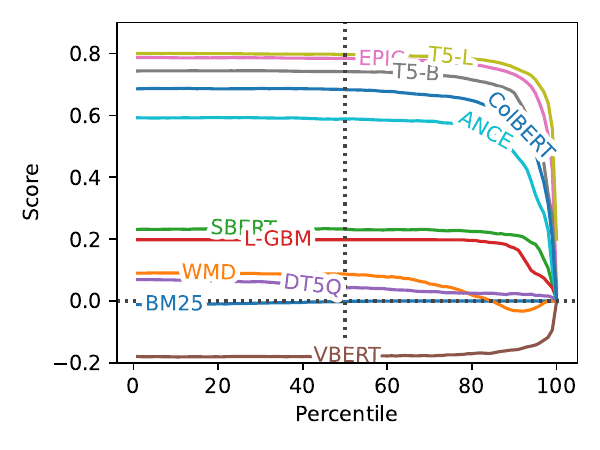} &
\includegraphics[scale=0.5]{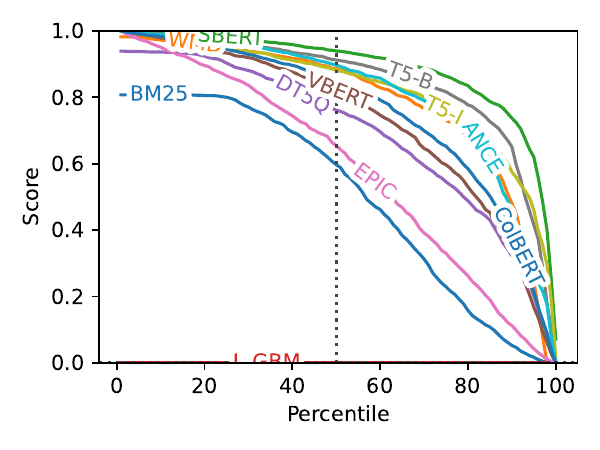} \\
\end{tabular}
\vspace{-2em}
\caption{Plots of scores for three representative probes when varying $
\delta$ to the specified percentile in TREC DL 2019. The vertical dashed line indicates the operational point of our experiments (the median value).}
\label{fig:deltas}
\end{figure*}

\smm{\textbf{Paraphrasing text can drastically change ranking scores.} In Table~\ref{tab:dtt}, we observe high scores across most models for the paraphrase probe. For BM25, this is because the document lengths differ to a substantial degree. Contextualized models---which one may expect to handle semantic equivalences like these well---assign substantially different scores for paraphrases up to \smb{94}\% of the time.}

\smm{To dig into specific stylistic differences that could explain the paraphrase discrepancies, we explore fluency, formality, simplicity, and neutrality. We find that fluency and formality have a greater effect than simplicity and neutrality. Most notably, EPIC and ColBERT prefer fluent text with scores of $+0.18$ to $+0.24$, while lexical models have low or insignificant differences. Meanwhile, EPIC and T5-L prefer formal text, while ColBERT and T5-B either prefer informal text or have no significant differences. Finally, the largest preferences observed for simple and neutral text are from BM25---which are likely a consequence of reduced document lengths.}

\smm{\textbf{Model behaviors vary considerably with succinctness.}
First, BM25 has a strong (+0.66) preference for the summaries in XSum, a moderate preference for summaries in CNN (+0.37), and no significant preference for Daily Mail. This suggests different standards among the various datasets, e.g., XSum (BBC) must use many of the same terms from the titles in the summaries, and provide long documents (reducing the score) that may not repeat terms from the title much. \smb{WMD also appears to be heavily affected by summaries, though in two of the three probes, there is insufficient evidence to claim significance.} The preference for summaries in XSum can be seen across all models except T5-B, which very slightly favors the full text.
Although most contextualized models prefer the full text for CNN and Daily Mail, VBERT prefers summaries for CNN ($+0.16$) while it prefers full text for Daily Mail ($-0.26$).
Such discrepancies warrant exploration in future work.}

\smm{\textbf{\smb{WMD, }T5, and ColBERT are biased towards factual answers.} From our factuality probes, we see that most models have little preference for factual passages. However, \smb{WMD}, both T5 variants, and ColBERT are biased towards answers that contain factually-correct information. \smb{For T5 and ColBERT}, this suggests that these models both learn some real-world information (likely in pre-training), and use this information as a signal when ranking. The larger size of T5-L appears to equip it with more knowledge, particularly about people, nationalities, and organizations. Curiously, although ColBERT exploits this information, the VBERT model (which uses the same base language model) does not appear to learn to use this information. \smb{For WMD, which doesn't have nearly the modeling capacity of T5 and ColBERT, the preference for factual information must be due to the fact that the word embeddings of the question are more similar to the word embeddings from the factual phrase than to those of the non-factual phrase. Although the contextualized language models should have the capacity to learn these trends and make similar decisions, this would be subject to such trends being present and distinguishable during fine-tuning. This suggests that using WMD over contextualized word embeddings may also improve the capacity of models to select factual answers.}}

\subsection{Effect of $\delta$}\label{sec:delta_effect}

\smm{Recall that $\delta$ defines the model-specific threshold at which a difference in ranking score is considered important. To test the importance of the selection of $\delta$, we test all probes while varying this parameter. Since the suitable values depend upon the range of scores for the particular ranker, we select $\delta$ by percentile among differences in the top 10 scoring passages of TREC DL 2019. Figure~\ref{fig:deltas} provides a representative sample of these plots. We find that for low percentiles (corresponding to settings where minute changes in score are considered important), the scores and rankings of systems can sometimes be unstable (e.g., see BM25 and DT5Q in (c)). This suggests that there are variations of the score distributions close to 0. However, we remind the reader that such differences are unlikely to have impactful changes in a real ranked list. We find that by the 50th percentile of $\delta$ (i.e., the value we use for our experiments), the rankings of the systems produced by \sys{} are generally stable. In most cases, the scores are stable as well, though in some cases drifting occurs (e.g., (c)).  With a large $\delta$, nearly no differences are considered important. \smb{In (c), we observe that L-GBM has no sensitivity to the paraphrases present in the probe, regardless of $\delta$.} These observations validate our technique for choosing $\delta$.}

\subsection{Effect of Model Training}\label{sec:random_init}

\smm{We observed that identical and similar base language models can differ in the behaviors they exhibit. To gain a better understanding of the origin of these differences, we probe 5 versions of the VBERT and EPIC models, each trained with different random seeds. We calculate the standard deviations of the performance over all the probes and report the average standard deviation for each probe type in Table~\ref{tab:random_init}. We find that among all probe types, MMPs are the most stable across random initializations and TMPs are the least stable. Curiously, the Stopword / punctuation removal TMP is the least stable probe across both models, with a stdev of 0.24 for VBERT and 0.27 for EPIC. In the case of VBERT, the probe score ranged from $-0.33$ to $+0.31$, highlighting that unexpected qualities can appear in models simply due to random variations in the training process. This is despite the fact that this probe is highly robust to the cutoff threshold on individual models (as seen in Figure~\ref{fig:deltas}(b)). Another probe with particularly high variance are the succinctness probe for VBERT using the CNN dataset, with a stdev of 0.23, and can either learn to prefer succinct ($+0.15$) or elaborative ($-0.42$) text, again due to the random initialization. These findings highlight that some biases can be introduced in the model training process randomly, rather than as a result of the pre-training process or model architecture.}

\begin{table}
\centering
\small
\begin{tabular}{lrr}
\toprule
Probes & VBERT Stdev. & EPIC Stdev. \\
\midrule
MMP & 3.5 &  3.6 \\
TMP & 11.2 & 17.1 \\
DTP & 9.5 & 8.9 \\
\bottomrule
\end{tabular}
\caption{Average standard deviations (square root of average variance) of 5 VBERT and EPIC models, by probe type.}
\label{tab:random_init}
\end{table}

\section{Related Work}

Pretrained contextualized language models are neural networks that are initially trained on language modeling objectives and are later fine-tuned on task-specific objectives~\cite{peters-etal-2018-deep}.
Well-known models include ELMo~\cite{peters-etal-2018-deep}, BERT~\cite{Devlin2019BERTPO}, and T5~\cite{Raffel2019ExploringTL}.
These models can effectively transfer signals to the task of ad-hoc retrieval, either by using the model directly (i.e., \textit{vanilla} or \textit{mono} models)~\cite{Nogueira2019PassageRW} or by using the outputs as features into a larger model~\cite{macavaney:sigir2019-cedr}. There has been a multitude of work in this area;
we refer the readers to~\citet{Lin2020PretrainedTF} for a comprehensive survey on these techniques.
We shed light on the mechanisms, strengths and weaknesses of this burgeoning body of work.

Diagnostic datasets, proposed by \citet{Rennings2019AnAA}, reformulate traditional ranking axioms---e.g., that documents with a higher term frequency should receive a higher ranking score~\cite{Fang2004AFS}---as empirical tests for analysing ranking models. \citeauthor{Rennings2019AnAA} studied neural ranking architectures that predate the rise of contextualized language models for ranking, and focused on just four axioms. \citet{Cmara2020DiagnosingBW} extended this work by adding five more previously-proposed ranking axioms (e.g., term proximity~\cite{Tao2007AnEO}, and word semantics~\cite{Fang2006SemanticTM}) and evaluating on a distilled BERT model. They found that the axioms are inadequate to explain the ranking effectiveness of their model. \smm{\citet{Vlske2021TowardsAE} examine the extent to which these axioms, when acting in concert, explain ranking model decisions.} Unlike these prior lines of work, we propose new probes that shed light onto possible sources of effectiveness, and test against current leading neural ranking architectures.

Although some insights about the effectiveness of contextualized language models for ranking have been gained using existing datasets \cite{Dai2019DeeperTU} and indirectly through various model architectures~\cite{Nogueira2019DocumentEB,Dai2019ContextAwareST,macavaney:sigir2020-epic,macavaney:sigir2019-cedr,Hofsttter2020InterpretableT,Khattab2020ColBERTEA}, they only provide circumstantial evidence. For instance, several works show how contextualized embedding similarity can be effective, but this does not imply that vanilla models utilize these signals for ranking. Rather than proposing new ranking models, in this work we analyze the effectiveness of existing models using controlled diagnostic probes, which allow us to gain insights into the particular behaviors and preferences of the ranking models.

Outside of the work in IR, others have developed techniques for investigating the behavior of contextualized language models in general. Although probing techniques~\cite{Tenney2019BERTRT} and attention analysis~\cite{Serrano2019IsAI} can be beneficial for understanding model capabilities, these techniques cannot help us characterize and quantify the behaviors of neural ranking models. CheckList~\cite{Ribeiro2020BeyondAB} and other challenge set techniques~\cite{McCoy2019RightFT} differ conceptually from our goals; we aim to characterize the behaviors to understand the qualities of ranking models, rather than provide additional measures of model quality.

\section{Conclusion}

We presented a new framework (\sys) for analyzing ranking models based on three probing strategies. By using probes from each strategy, we demonstrated that a variety of insights can be gained about the behaviors of recently-proposed ranking models, such as those based on BERT and T5. Our analysis is, to date, the most extensive analysis of the behaviors of neural ranking models, and sheds light on several unexpected model behaviors. For instance, adding non-relevant text can increase a document's ranking score, even though the models are largely not biased towards longer documents. We also see that the same base language model used with a different ranking architecture can yield different behaviors, such as higher sensitivity to shuffling a document's text. \smm{We also find that some models learn to utilize real-world knowledge in the ranking process. Finally, we observe that some strong biases can appear simply by chance during the training process. This motivates future investigations on approaches for stabilizing training processes and avoiding the introduction of unwanted biases.}

\bibliography{biblio}

\begin{thebibliography}{61}
\expandafter\ifx\csname natexlab\endcsname\relax\def\natexlab#1{#1}\fi

\bibitem[{Alleman et~al.(2021)Alleman, Mamou, Rio, Tang, Kim, and
  Chung}]{Alleman2021SyntacticPR}
Matteo Alleman, J.~Mamou, M.~D. Rio, Hanlin Tang, Yoon Kim, and SueYeon Chung.
  2021.
\newblock Syntactic perturbations reveal representational correlates of
  hierarchical phrase structure in pretrained language models.
\newblock \emph{ArXiv}, abs/2104.07578.

\bibitem[{C{\^a}mara and Hauff(2020)}]{Cmara2020DiagnosingBW}
Arthur C{\^a}mara and Claudia Hauff. 2020.
\newblock Diagnosing {BERT} with retrieval heuristics.
\newblock In \emph{ECIR}.

\bibitem[{Campos et~al.(2016)Campos, Nguyen, Rosenberg, Song, Gao, Tiwary,
  Majumder, Deng, and Mitra}]{Campos2016MSMA}
Daniel~Fernando Campos, T.~Nguyen, M.~Rosenberg, Xia Song, Jianfeng Gao,
  Saurabh Tiwary, Rangan Majumder, L.~Deng, and Bhaskar Mitra. 2016.
\newblock {MS MARCO}: A human generated machine reading comprehension dataset.
\newblock \emph{ArXiv}, abs/1611.09268.

\bibitem[{Craswell et~al.(2019)Craswell, Mitra, Yilmaz, Campos, and
  Voorhees}]{craswell2020overview}
Nick Craswell, Bhaskar Mitra, Emine Yilmaz, Daniel Campos, and Ellen~M
  Voorhees. 2019.
\newblock Overview of the {TREC} 2019 deep learning track.
\newblock In \emph{TREC}.

\bibitem[{Dai and Callan(2019{\natexlab{a}})}]{Dai2019ContextAwareST}
Zhuyun Dai and J.~Callan. 2019{\natexlab{a}}.
\newblock Context-aware sentence/passage term importance estimation for first
  stage retrieval.
\newblock \emph{ArXiv}, abs/1910.10687.

\bibitem[{Dai and Callan(2019{\natexlab{b}})}]{Dai2019DeeperTU}
Zhuyun Dai and J.~Callan. 2019{\natexlab{b}}.
\newblock Deeper text understanding for ir with contextual neural language
  modeling.
\newblock \emph{SIGIR}.

\bibitem[{Dalton et~al.(2019)Dalton, Xiong, and Callan}]{dalton2019cast}
Jeffrey Dalton, Chenyan Xiong, and Jamie Callan. 2019.
\newblock {CAsT} 2019: The conversational assistance track overview.
\newblock In \emph{TREC}.

\bibitem[{Devlin et~al.(2019)Devlin, Chang, Lee, and
  Toutanova}]{Devlin2019BERTPO}
J.~Devlin, Ming-Wei Chang, Kenton Lee, and Kristina Toutanova. 2019.
\newblock {BERT}: Pre-training of deep bidirectional transformers for language
  understanding.
\newblock In \emph{NAACL-HLT}.

\bibitem[{Fang et~al.(2004)Fang, Tao, and Zhai}]{Fang2004AFS}
Hui Fang, T.~Tao, and ChengXiang Zhai. 2004.
\newblock A formal study of information retrieval heuristics.
\newblock In \emph{SIGIR}.

\bibitem[{Fang et~al.(2011)Fang, Tao, and Zhai}]{Fang2011DiagnosticEO}
Hui Fang, T.~Tao, and ChengXiang Zhai. 2011.
\newblock Diagnostic evaluation of information retrieval models.
\newblock \emph{ACM Trans. Inf. Syst.}, 29:7:1--7:42.

\bibitem[{Fang and Zhai(2006)}]{Fang2006SemanticTM}
Hui Fang and ChengXiang Zhai. 2006.
\newblock Semantic term matching in axiomatic approaches to information
  retrieval.
\newblock In \emph{SIGIR '06}.

\bibitem[{Feldman(2020)}]{feldman_2020}
Sergey Feldman. 2020.
\newblock \href
  {https://medium.com/ai2-blog/building-a-better-search-engine-for-semantic-scholar-ea23a0b661e7}
  {Building a better search engine for semantic scholar}.
\newblock Blog post.

\bibitem[{Hashemi et~al.(2020)Hashemi, Aliannejadi, Zamani, and
  Croft}]{Hashemi2020Antique}
Helia Hashemi, Mohammad Aliannejadi, Hamed Zamani, and Bruce Croft. 2020.
\newblock {ANTIQUE}: A non-factoid question answering benchmark.
\newblock In \emph{ECIR}.

\bibitem[{Heafield et~al.(2013)Heafield, Pouzyrevsky, Clark, and
  Koehn}]{heafield-etal-2013-scalable}
Kenneth Heafield, Ivan Pouzyrevsky, Jonathan~H. Clark, and Philipp Koehn. 2013.
\newblock Scalable modified {K}neser-{N}ey language model estimation.
\newblock In \emph{ACL}.

\bibitem[{Hofst{\"a}tter et~al.(2020)Hofst{\"a}tter, Zlabinger, and
  Hanbury}]{Hofsttter2020InterpretableT}
Sebastian Hofst{\"a}tter, Markus Zlabinger, and A.~Hanbury. 2020.
\newblock Interpretable and time-budget-constrained contextualization for
  re-ranking.
\newblock In \emph{ECAI}.

\bibitem[{Honnibal and Montani(2017)}]{spacy2}
Matthew Honnibal and Ines Montani. 2017.
\newblock {spaCy 2}: Natural language understanding with {B}loom embeddings,
  convolutional neural networks and incremental parsing.
\newblock To appear.

\bibitem[{Ke et~al.(2017)Ke, Meng, Finley, Wang, Chen, Ma, Ye, and
  Liu}]{Ke2017LightGBMAH}
Guolin Ke, Q.~Meng, Thomas Finley, Taifeng Wang, Wei Chen, Weidong Ma, Qiwei
  Ye, and T.~Liu. 2017.
\newblock {LightGBM}: A highly efficient gradient boosting decision tree.
\newblock In \emph{NIPS}.

\bibitem[{Khattab and Zaharia(2020)}]{Khattab2020ColBERTEA}
O.~Khattab and M.~Zaharia. 2020.
\newblock {ColBERT}: Efficient and effective passage search via contextualized
  late interaction over {BERT}.
\newblock In \emph{SIGIR}.

\bibitem[{Kusner et~al.(2015)Kusner, Sun, Kolkin, and
  Weinberger}]{Kusner2015FromWE}
Matt~J. Kusner, Yu~Sun, Nicholas~I. Kolkin, and Kilian~Q. Weinberger. 2015.
\newblock From word embeddings to document distances.
\newblock In \emph{ICML}.

\bibitem[{Kwiatkowski et~al.(2019)Kwiatkowski, Palomaki, Redfield, Collins,
  Parikh, Alberti, Epstein, Polosukhin, Kelcey, Devlin, Lee, Toutanova, Jones,
  Chang, Dai, Uszkoreit, Le, and Petrov}]{Kwiatkowski2019NQ}
Tom Kwiatkowski, Jennimaria Palomaki, Olivia Redfield, Michael Collins, Ankur
  Parikh, Chris Alberti, Danielle Epstein, Illia Polosukhin, Matthew Kelcey,
  Jacob Devlin, Kenton Lee, Kristina~N. Toutanova, Llion Jones, Ming-Wei Chang,
  Andrew Dai, Jakob Uszkoreit, Quoc Le, and Slav Petrov. 2019.
\newblock {Natural Questions}: a benchmark for question answering research.
\newblock \emph{TACL}.

\bibitem[{Li et~al.(2020)Li, Yates, MacAvaney, He, and
  Sun}]{li:arxiv2020-parade}
Canjia Li, Andrew Yates, Sean MacAvaney, Ben He, and Yingfei Sun. 2020.
\newblock Parade: Passage representation aggregation for document reranking.
\newblock \emph{arXiv}, abs/2008.09093.

\bibitem[{Lin et~al.(2020)Lin, Nogueira, and Yates}]{Lin2020PretrainedTF}
Jimmy Lin, Rodrigo Nogueira, and A.~Yates. 2020.
\newblock Pretrained transformers for text ranking: {BERT} and beyond.
\newblock \emph{ArXiv}, abs/2010.06467.

\bibitem[{Litschko et~al.(2021)Litschko, Vuli'c, Ponzetto, and
  Glavavs}]{Litschko2021EvaluatingMT}
Robert Litschko, Ivan Vuli'c, Simone~Paolo Ponzetto, and Goran Glavavs. 2021.
\newblock Evaluating multilingual text encoders for unsupervised cross-lingual
  retrieval.
\newblock In \emph{ECIR}.

\bibitem[{Liu et~al.(2019{\natexlab{a}})Liu, Gardner, Belinkov, Peters, and
  Smith}]{Liu2019LinguisticKA}
Nelson~F. Liu, Matt Gardner, Yonatan Belinkov, Matthew~E. Peters, and Noah~A.
  Smith. 2019{\natexlab{a}}.
\newblock Linguistic knowledge and transferability of contextual
  representations.
\newblock In \emph{NAACL-HLT}.

\bibitem[{Liu et~al.(2019{\natexlab{b}})Liu, Ott, Goyal, Du, Joshi, Chen, Levy,
  Lewis, Zettlemoyer, and Stoyanov}]{Liu2019RoBERTaAR}
Yinhan Liu, Myle Ott, Naman Goyal, Jingfei Du, Mandar Joshi, Danqi Chen, Omer
  Levy, M.~Lewis, Luke Zettlemoyer, and Veselin Stoyanov. 2019{\natexlab{b}}.
\newblock {RoBERTa}: A robustly optimized {BERT} pretraining approach.
\newblock \emph{ArXiv}, abs/1907.11692.

\bibitem[{Loureiro et~al.(2020)Loureiro, Rezaee, Pilehvar, and
  Camacho-Collados}]{Loureiro2020LanguageMA}
D.~Loureiro, Kiamehr Rezaee, Mohammad~Taher Pilehvar, and Jos{\'e}
  Camacho-Collados. 2020.
\newblock Language models and word sense disambiguation: An overview and
  analysis.
\newblock \emph{ArXiv}, abs/2008.11608.

\bibitem[{MacAvaney(2020)}]{macavaney:wsdm2020-onir}
Sean MacAvaney. 2020.
\newblock {OpenNIR}: A complete neural ad-hoc ranking pipeline.
\newblock In \emph{WSDM}.

\bibitem[{MacAvaney et~al.(2020)MacAvaney, Nardini, Perego, Tonellotto,
  Goharian, and Frieder}]{macavaney:sigir2020-epic}
Sean MacAvaney, Franco~Maria Nardini, Raffaele Perego, Nicola Tonellotto, Nazli
  Goharian, and Ophir Frieder. 2020.
\newblock Expansion via prediction of importance with contextualization.
\newblock In \emph{SIGIR}.

\bibitem[{MacAvaney et~al.(2019{\natexlab{a}})MacAvaney, Yates, Cohan, and
  Goharian}]{macavaney:sigir2019-cedr}
Sean MacAvaney, Andrew Yates, Arman Cohan, and Nazli Goharian.
  2019{\natexlab{a}}.
\newblock {CEDR}: Contextualized embeddings for document ranking.
\newblock In \emph{SIGIR}.

\bibitem[{MacAvaney et~al.(2021)MacAvaney, Yates, Feldman, Downey, Cohan, and
  Goharian}]{macavaney:sigir2021-irds}
Sean MacAvaney, Andrew Yates, Sergey Feldman, Doug Downey, Arman Cohan, and
  Nazli Goharian. 2021.
\newblock Simplified data wrangling with ir\_datasets.
\newblock In \emph{SIGIR}.

\bibitem[{MacAvaney et~al.(2019{\natexlab{b}})MacAvaney, Yates, Hui, and
  Frieder}]{macavaney:sigir2019-nyt}
Sean MacAvaney, Andrew Yates, Kai Hui, and Ophir Frieder. 2019{\natexlab{b}}.
\newblock Content-based weak supervision for ad-hoc re-ranking.
\newblock In \emph{SIGIR}.

\bibitem[{Macdonald et~al.(2021)Macdonald, Tonellotto, MacAvaney, and
  Ounis}]{MacDonald2020DeclarativeEI}
Craig Macdonald, Nicola Tonellotto, Sean MacAvaney, and Iadh Ounis. 2021.
\newblock {PyTerrier}: Declarative experimentation in python from {BM25} to
  dense retrieval.
\newblock In \emph{CIKM}.

\bibitem[{McCoy et~al.(2019)McCoy, Pavlick, and Linzen}]{McCoy2019RightFT}
R.~Thomas McCoy, Ellie Pavlick, and Tal Linzen. 2019.
\newblock Right for the wrong reasons: Diagnosing syntactic heuristics in
  natural language inference.
\newblock In \emph{ACL}.

\bibitem[{Napoles et~al.(2017)Napoles, Sakaguchi, and
  Tetreault}]{napoles-sakaguchi-tetreault:2017:EACLshort}
Courtney Napoles, Keisuke Sakaguchi, and Joel Tetreault. 2017.
\newblock {JFLEG}: A fluency corpus and benchmark for grammatical error
  correction.
\newblock In \emph{EACL}.

\bibitem[{Narayan et~al.(2018)Narayan, Cohen, and Lapata}]{xsum-emnlp}
Shashi Narayan, Shay~B. Cohen, and Mirella Lapata. 2018.
\newblock Don't give me the details, just the summary! {T}opic-aware
  convolutional neural networks for extreme summarization.
\newblock In \emph{EMNLP}.

\bibitem[{Nogueira and Cho(2019)}]{Nogueira2019PassageRW}
Rodrigo Nogueira and Kyunghyun Cho. 2019.
\newblock Passage re-ranking with {BERT}.
\newblock \emph{ArXiv}, abs/1901.04085.

\bibitem[{Nogueira et~al.(2020)Nogueira, Jiang, and
  Lin}]{Nogueira2020DocumentRW}
Rodrigo Nogueira, Zhiying Jiang, and Jimmy Lin. 2020.
\newblock Document ranking with a pretrained sequence-to-sequence model.
\newblock \emph{ArXiv}, abs/2003.06713.

\bibitem[{Nogueira and Lin(2019)}]{nogueiradoc2query}
Rodrigo Nogueira and Jimmy Lin. 2019.
\newblock \href
  {https://cs.uwaterloo.ca/~jimmylin/publications/Nogueira_Lin_2019_docTTTTTquery-v2.pdf}
  {From doc2query to {docTTTTTquery}}.
\newblock Self-published.

\bibitem[{Nogueira et~al.(2019)Nogueira, Yang, Lin, and
  Cho}]{Nogueira2019DocumentEB}
Rodrigo Nogueira, Wei Yang, Jimmy Lin, and Kyunghyun Cho. 2019.
\newblock Document expansion by query prediction.
\newblock \emph{ArXiv}, abs/1904.08375.

\bibitem[{Ounis et~al.(2006)Ounis, Amati, Plachouras, He, Macdonald, and
  Lioma}]{ounis06terrier-osir}
I.~Ounis, G.~Amati, V.~Plachouras, B.~He, C.~Macdonald, and C.~Lioma. 2006.
\newblock Terrier: A high performance and scalable information retrieval
  platform.
\newblock In \emph{Proceedings of ACM SIGIR'06 Workshop on Open Source
  Information Retrieval (OSIR 2006)}.

\bibitem[{Pennington et~al.(2014)Pennington, Socher, and
  Manning}]{Pennington2014GloVeGV}
Jeffrey Pennington, Richard Socher, and Christopher~D. Manning. 2014.
\newblock {GloVe}: Global vectors for word representation.
\newblock In \emph{EMNLP}.

\bibitem[{Peters et~al.(2018)Peters, Neumann, Iyyer, Gardner, Clark, Lee, and
  Zettlemoyer}]{peters-etal-2018-deep}
Matthew Peters, Mark Neumann, Mohit Iyyer, Matt Gardner, Christopher Clark,
  Kenton Lee, and Luke Zettlemoyer. 2018.
\newblock Deep contextualized word representations.
\newblock In \emph{NAACL-HLT}.

\bibitem[{Pryzant et~al.(2020)Pryzant, Martinez, Dass, Kurohashi, Jurafsky, and
  Yang}]{Pryzant2020AutomaticallyNS}
Reid Pryzant, Richard~Diehl Martinez, Nathan Dass, S.~Kurohashi, Dan Jurafsky,
  and Diyi Yang. 2020.
\newblock Automatically neutralizing subjective bias in text.
\newblock In \emph{AAAI}.

\bibitem[{Raffel et~al.(2020)Raffel, Shazeer, Roberts, Lee, Narang, Matena,
  Zhou, Li, and Liu}]{Raffel2019ExploringTL}
Colin Raffel, Noam Shazeer, Adam Roberts, Katherine Lee, Sharan Narang, Michael
  Matena, Yanqi Zhou, Wei Li, and Peter~J. Liu. 2020.
\newblock Exploring the limits of transfer learning with a unified text-to-text
  transformer.
\newblock \emph{Journal of Machine Learning Research}, 21(140).

\bibitem[{Rao and Tetreault(2018)}]{Rao2018DearSO}
Sudha Rao and J.~Tetreault. 2018.
\newblock Dear sir or madam, may {I} introduce the {YAFC} corpus: Corpus,
  benchmarks and metrics for formality style transfer.
\newblock In \emph{NAACL-HLT}.

\bibitem[{Rehurek and Sojka(2011)}]{rehurek2011gensim}
Radim Rehurek and Petr Sojka. 2011.
\newblock Gensim--{Python} framework for vector space modelling.
\newblock \emph{NLP Centre, Faculty of Informatics, Masaryk University, Brno,
  Czech Republic}, 3(2).

\bibitem[{Reimers and Gurevych(2019)}]{reimers-2019-sentence-bert}
Nils Reimers and Iryna Gurevych. 2019.
\newblock Sentence-{BERT}: Sentence embeddings using siamese {BERT}-networks.
\newblock In \emph{EMNLP}. Association for Computational Linguistics.

\bibitem[{Rennings et~al.(2019)Rennings, Moraes, and Hauff}]{Rennings2019AnAA}
D.~Rennings, Felipe Moraes, and C.~Hauff. 2019.
\newblock An axiomatic approach to diagnosing neural {IR} models.
\newblock In \emph{ECIR}.

\bibitem[{Ribeiro et~al.(2020)Ribeiro, Wu, Guestrin, and
  Singh}]{Ribeiro2020BeyondAB}
Marco~T{\'u}lio Ribeiro, Tongshuang Wu, Carlos Guestrin, and Sameer Singh.
  2020.
\newblock Beyond accuracy: Behavioral testing of {NLP} models with checklist.
\newblock In \emph{ACL}.

\bibitem[{Rogers et~al.(2020)Rogers, Kovaleva, and Rumshisky}]{Rogers2020API}
Anna Rogers, O.~Kovaleva, and Anna Rumshisky. 2020.
\newblock A primer in {BERTology}: What we know about how {BERT} works.
\newblock \emph{TACL}.

\bibitem[{See et~al.(2017)See, Liu, and Manning}]{See2017GetTT}
A.~See, Peter~J. Liu, and Christopher~D. Manning. 2017.
\newblock Get to the point: Summarization with pointer-generator networks.
\newblock In \emph{ACL}.

\bibitem[{Serrano and Smith(2019)}]{Serrano2019IsAI}
Sofia Serrano and Noah~A. Smith. 2019.
\newblock Is attention interpretable?
\newblock In \emph{ACL}.

\bibitem[{Sinha et~al.(2021)Sinha, Jia, Hupkes, Pineau, Williams, and
  Kiela}]{Sinha2021MaskedLM}
Koustuv Sinha, Robin Jia, Dieuwke Hupkes, J.~Pineau, Adina Williams, and Douwe
  Kiela. 2021.
\newblock Masked language modeling and the distributional hypothesis: Order
  word matters pre-training for little.
\newblock \emph{ArXiv}, abs/2104.06644.

\bibitem[{Tao and Zhai(2007)}]{Tao2007AnEO}
T.~Tao and ChengXiang Zhai. 2007.
\newblock An exploration of proximity measures in information retrieval.
\newblock In \emph{SIGIR}.

\bibitem[{Tenney et~al.(2019)Tenney, Das, and Pavlick}]{Tenney2019BERTRT}
Ian Tenney, Dipanjan Das, and Ellie Pavlick. 2019.
\newblock {BERT} rediscovers the classical {NLP} pipeline.
\newblock In \emph{ACL}.

\bibitem[{V{\"o}lske et~al.(2021)V{\"o}lske, Bondarenko, Fr{\"o}be, Hagen,
  Stein, Singh, and Anand}]{Vlske2021TowardsAE}
Michael V{\"o}lske, A.~Bondarenko, Maik Fr{\"o}be, Matthias Hagen, Benno Stein,
  Jaspreet Singh, and Avishek Anand. 2021.
\newblock Towards axiomatic explanations for neural ranking models.
\newblock \emph{ArXiv}, abs/2106.08019.

\bibitem[{Wolf et~al.(2019)Wolf, Debut, Sanh, Chaumond, Delangue, Moi, Cistac,
  Rault, Louf, Funtowicz, and Brew}]{Wolf2019HuggingFacesTS}
Thomas Wolf, Lysandre Debut, Victor Sanh, Julien Chaumond, Clement Delangue,
  Anthony Moi, Pierric Cistac, Tim Rault, R'emi Louf, Morgan Funtowicz, and
  Jamie Brew. 2019.
\newblock {HuggingFace}'s {Transformers}: State-of-the-art natural language
  processing.
\newblock \emph{ArXiv}, abs/1910.03771.

\bibitem[{Xia et~al.(2020)Xia, Wu, and Durme}]{Xia2020WhichA}
Patrick Xia, Shijie Wu, and B.~Van Durme. 2020.
\newblock Which *{BERT}? a survey organizing contextualized encoders.
\newblock In \emph{EMNLP}.

\bibitem[{Xiong et~al.(2021)Xiong, Xiong, Li, Tang, Liu, Bennett, Ahmed, and
  Overwijk}]{Xiong2021ApproximateNN}
Lee Xiong, Chenyan Xiong, Ye~Li, Kwok-Fung Tang, Jialin Liu, Paul Bennett,
  Junaid Ahmed, and Arnold Overwijk. 2021.
\newblock Approximate nearest neighbor negative contrastive learning for dense
  text retrieval.
\newblock \emph{ArXiv}, abs/2007.00808.

\bibitem[{Xu et~al.(2016)Xu, Napoles, Pavlick, Chen, and
  Callison-Burch}]{xu-etal-2016-optimizing}
Wei Xu, Courtney Napoles, Ellie Pavlick, Quanze Chen, and Chris Callison-Burch.
  2016.
\newblock Optimizing statistical machine translation for text simplification.
\newblock \emph{TACL}, 4.

\bibitem[{Yang et~al.(2018)Yang, Fang, and Lin}]{Yang2018AnseriniRR}
Peilin Yang, Hui Fang, and Jimmy Lin. 2018.
\newblock Anserini: Reproducible ranking baselines using {Lucene}.
\newblock \emph{J. Data and Information Quality}, 10.

\end{thebibliography}
\bibliographystyle{acl_natbib}

\end{document}